\documentclass{article}
\usepackage{PRIMEarxiv}

\usepackage[utf8]{inputenc} 
\usepackage[T1]{fontenc}    
\usepackage{url}            
\usepackage{booktabs}       
\usepackage{amsfonts}       
\usepackage{nicefrac}       
\usepackage{microtype}      
\usepackage{lipsum}
\usepackage{fancyhdr}       
\usepackage{graphicx}       

\usepackage[backend=biber, style=authoryear-comp, natbib=true, maxcitenames=1, uniquelist=false, maxnames=1, uniquename=false]{biblatex}
\usepackage{placeins}
\usepackage{rotating}
\usepackage{afterpage}
\usepackage{lipsum}
\usepackage{lscape} 
\usepackage{tablefootnote}
\usepackage{nicematrix}
\usepackage{booktabs}
\usepackage{multirow, makecell}
\usepackage{enumitem}
\usepackage{amsmath}

\usepackage[acronym]{glossaries}
\usepackage{varioref} 
\usepackage{hyperref}
\usepackage{cleveref}
\usepackage[font=small,labelfont=bf]{caption}
\usepackage{subcaption}
\usepackage{graphicx}
\graphicspath{ {figures/} }

\newacronym{sar}{SAR}{Synthetic Aperture Radar}
\newacronym{insar}{InSAR}{Interferometric Synthetic Aperture Radar}
\newacronym{dem}{DEM}{Digital Elevation Model}
\newacronym{lidar}{LiDAR}{Light Detection and Ranging}
\newacronym{vhr}{VHR}{Very High-Resolution}

\newacronym{mb}{MB}{Mass Balance}

\newacronym{rgi}{RGI}{Randolph Glacier Inventory}
\newacronym{sgi}{SGI}{Swiss Glacier Inventory}
\newacronym{gamdam}{GAMDAM}{Glacier Area Mapping for Discharge in Asian Mountains}
\newacronym{glims}{GLIMS}{Global Land Ice Measurements from Space}
\newacronym{glamos}{GLAMOS}{Glacier Monitoring in Switzerland}
\newacronym{hma}{HMA}{High-Mountain Asia}
\newacronym{ipa}{IPA}{International Permafrost Association}
\newacronym{iou}{IOU}{Intersection Over Union}

\newacronym{ecv}{ECV}{Essential Climate Variable}
\newacronym{eo}{EO}{Earth Observation}
\newacronym{swir}{SWIR}{Short-Wave InfraRed}
\newacronym{nir}{NIR}{Near-InfraRed}
\newacronym{ndvi}{NDVI}{Normalized Difference Vegetation Index}
\newacronym{ndwi}{NDWI}{Normalized Difference Water Index}
\newacronym{mndwi}{MNDWI}{Modified Normalized Difference Water Index}
\newacronym{ndsi}{NDSI}{Normalized Difference Snow Index}
\newacronym{savi}{SAVI}{Soil Adjusted Vegetation Index}
\newacronym{nbr}{NBR}{Normalized Burn Ratio}
\newacronym{gsd}{GSD}{Ground Sample Distance}

\newacronym[\glslongpluralkey=Regions of Interest]{roi}{ROI}{Region of Interest}
\newacronym{dl}{DL}{Deep Learning}
\newacronym{ml}{ML}{Machine Learning}
\newacronym{cnn}{CNN}{Convolutional Neural Network}
\newacronym{mlp}{MLP}{Multi-Layer Perceptron}
\newacronym{obia}{OBIA}{Object-Based Image Analysis}

\title{Multi-Sensor Deep Learning for Glacier Mapping}

\begin{document}
\maketitle
\vspace{-7em}
\begin{flushleft}
\textbf{Codruț-Andrei Diaconu\textsuperscript{1, 2. *}},
\textbf{Konrad Heidler\textsuperscript{2}},
\textbf{Jonathan L. Bamber\textsuperscript{2, 3}},
\textbf{Harry Zekollari\textsuperscript{4, 5. 6}}
\\
\bigskip
\textsuperscript{1}Earth Observation Center, German Aerospace Center (DLR), Germany
\\
\textsuperscript{2}School of Engineering and Design, Technical University of Munich, Germany
\\
\textsuperscript{3}Bristol Glaciology Centre, University of Bristol, United Kingdom
\\
\textsuperscript{4}Department of Water and Climate, Vrije Universiteit Brussel, Belgium
\\
\textsuperscript{5}Laboratory of Hydraulics, Hydrology and Glaciology (VAW), ETH Zurich, Switzerland
\\
\textsuperscript{6}Laboratoire de Glaciologie, Université Libre de Bruxelles, Belgium
\\
\bigskip
\textit{\small{\textsuperscript{*}codrut-andrei.diaconu@dlr.de}}
\end{flushleft}
\vspace{2em}

\begin{abstract}
\let\thefootnote\relax\footnotetext{This article will be a chapter of the book \textit{Deep Learning for Multi-Sensor Earth Observation}, to be published by Elsevier.}

The more than 200,000 glaciers outside the ice sheets play a crucial role in our society by influencing sea-level rise, water resource management, natural hazards, biodiversity, and tourism. However, only a fraction of these glaciers benefit from consistent and detailed in-situ observations that allow for assessing their status and changes over time. This limitation can, in part, be overcome by relying on satellite-based Earth Observation techniques. Satellite-based glacier mapping applications have historically mainly relied on manual and semi-automatic detection methods, while recently, a fast and notable transition to deep learning techniques has started.

This chapter reviews how combining multi-sensor remote sensing data and deep learning allows us to better delineate (i.e. map) glaciers and detect their temporal changes. We explain how relying on deep learning multi-sensor frameworks to map glaciers benefits from the extensive availability of regional and global glacier inventories. We also analyse the rationale behind glacier mapping, the benefits of deep learning methodologies, and the inherent challenges in integrating multi-sensor earth observation data with deep learning algorithms. 

While our review aims to provide a broad overview of glacier mapping efforts, we highlight a few setups where deep learning multi-sensor remote sensing applications have a considerable potential added value. This includes applications for debris-covered and rock glaciers that are visually difficult to distinguish from surroundings and for calving glaciers that are in contact with the ocean. These specific cases are illustrated through a series of visual imageries, highlighting some significant advantages and challenges when detecting glacier changes, including dealing with seasonal snow cover, changing debris coverage, and distinguishing glacier fronts from the surrounding sea ice.
\end{abstract}


\section{Introduction}
There are ca. 275,000 glaciers around the world \citep{rgi_consortium_randolph_2023}, which act as important contributors to sea-level rise \citep{hugonnet_accelerated_2021, edwards_projected_2021}, water resources \citep{immerzeel_importance_2020}, triggers of natural hazards \citep{veh_less_2023}, biodiversity regulators \citep{bosson_future_2023}, and touristic attractions \citep{salim_glacier_2023}. Around 0.1 percent of these glaciers have a reasonably long record of in-situ observations of mass change \citep{zemp_historically_2015, wgms}. Therefore, satellite-based \gls{eo} is the only feasible way to obtain representative sampling across the broad latitudinal and altitudinal range occupied by glaciers. Glacier-specific observations are essential because i) individual glaciers can respond to the climate in complex ways that depend on their geometry, morphology and other boundary conditions \citep{brun_heterogeneous_2019} while ii) as an ensemble, glaciers are valuable and unique indicators of integrated climate change over multi-decadal to centennial timescales \citep{marzeion_limited_2018}. Numerous satellite missions and sensors can, and have been, used to measure glacier changes, often with conflicting results \citep{gardner_reconciled_2013, zemp_global_2019, wouters_global_2019}. 

\gls{ml}, in particular multi-modal approaches based on \gls{dl}, offer tremendous potential. These methods can extract information, patterns and trends that have proven to be challenging to obtain through "conventional" approaches. While \gls{dl} methods in \gls{eo}, in particular for land classification, are well-developed fields with a rich history, application of \gls{dl} methods in glacier (change) detection only recently started to emerge in the literature.  Despite being a recent field of research, the application of \gls{dl} for glacier mapping has rapidly progressed. We believe this rapid evolution is, in part, due to the following reasons: i) the glacier mapping task is mainly a segmentation task, and, as a consequence, many Computer Vision methods are directly applicable  \citep{xie_evaluating_2021}; ii) there are many existing regional glaciers inventories, including a global one \textemdash \gls{rgi} \citep{rgi_consortium_randolph_2012, rgi_consortium_randolph_2023} \textemdash which provide the necessary training labels, and iii) the level of complexity of glacier mapping is less than for other glacier-related problems (e.g. see \Cref{sec:discussion-glacier-evolution} where we also shed light on \gls{dl} developments related to glacier evolution modelling) where the input-output relations are more complicated, often requiring additional physical constraints, and typically suffering from a lack of extensive, high-quality training data.

Here, we discuss how recent glacier mapping efforts build on segmentation algorithms, emphasizing its particular challenges. We specifically focus on the glaciers outside the ice sheets while also including ice-sheet outlet glaciers.

\subsection*{Motivation}

A review of Deep Learning applications for cryospheric studies, including, e.g. glaciers, ice sheets, permafrost, and snow, is provided by \citet{campsvalls_review_2021}. Their review provides a broad perspective on the field, focusing on a few selected studies. Here, we instead focus on a single application area, i.e. glacier mapping, which is rapidly growing, thereby developing maturity from a \gls{dl} perspective. 

To our knowledge, this is the first study that provides an overview of the glacier mapping literature based on \gls{dl}. Here, we describe some of the significant studies published so far and summarize which data sources have been used, highlighting that glacier mapping is, to a large extent, a multi-modal task.

\subsection*{Structure} In \Cref{sec:intro-gl-mapping}, we provide an introduction for glacier mapping: we explain why this is an essential problem in cryospheric studies, then discuss the benefits of relying on \gls{dl} and finally provide an overview of the associated challenges. \Cref{sec:data-modalities} is dedicated to the data modalities used in the studies included in our literature review  (\Cref{sec:literature-overview}), which covers two major topics in glacier mapping, detecting i) the full extent of a glacier (\Cref{sec:glacier-extent-mapping}) and ii) calving fronts (\Cref{sec:cf-detection}). Next, we provide a discussion in \Cref{sec:discussion} followed by a summary (\Cref{sec:summary}). Additionally, in \Cref{sec:resources} we include a list of resources (mainly databases) that could be exploited with \gls{dl}.

\section{Glacier Mapping with Deep Learning}

\label{sec:intro-gl-mapping}

\subsection*{Glacier Mapping}
To study various properties of glaciers, e.g. area, hypsometry, we first need to know where they are located. Various regional glacier inventories have been produced, leading to the first global glacier inventory, the \gls{rgi} \citep{rgi_consortium_randolph_2012} that was initially derived from \gls{glims}, which is a multi-temporal glacier database. The \gls{rgi} was initially developed as part of the Fifth Assessment Report of the Intergovernmental Panel on Climate Change \citep{stocker_ipcc_2013} and was designed to be a snapshot of all the glaciers in the world at the beginning of the 21st century. There are, however, significant variations among the subregions, as \gls{rgi} is a compilation of regional inventories from various sources, most of which are based on satellite imagery from 1999–2010. In the latest version \citep{rgi_consortium_randolph_2023}, the \gls{rgi} contains close to 275,000 glaciers, with a minimum area of 0.01 km$^2$, covering a total surface of $\sim$707,000 km$^2$, with an estimated $\pm$5\% error. Using \gls{rgi}, we can derive the glacier area distribution with elevation, an essential input in many glacier evolution models \citep{marzeion_partitioning_2020, zekollari_ice-dynamical_2022}, thus making \gls{rgi} a critical dataset, being constantly enhanced.

Glaciers have been defined by \citet{bojinski}, under the Global Climate Observing System, as an Essential Climate Variable (ECV), given that glaciers are sensitive and unique indicators of climate change \citep{hock_glaciers_2021}. Glacier area, elevation change and mass change/balance are listed as ECV products. Various studies have shown a significant glacier retreat in different parts of the world. Such studies usually imply (re)creating the glacier outlines at two different points in time and analyzing the differences. For instance, \citet{paul_glacier_2020} rebuilt a glacier inventory for the European Alps and compared the total glaciated area with the previous inventory (from \gls{rgi}), showing a regional area change rate of ca. -1.2\% per year from 2003 to 2015. 
A more recent study by \citet{ali_glacier_2023} used \gls{obia} to produce the outlines of ca. 2200 glaciers in the Arctic from four different regions at three time periods: 1985–1989, 2000–2002, and 2019–2021. Their results show an accelerated loss in the second period, e.g. from -36.1 km$^2$ yr$^{-1}$ to -41 km$^2$ yr$^{-1}$ for Novaya Zemlya and from -3.7 km$^2$ yr$^{-1}$ to -8 km$^2$ yr$^{-1}$ in Kenai. They also show that 73 glaciers have entirely disappeared.  When imposing the same ice divides on the glacier outlines, each glacier can be individually tracked over time, allowing to capture the heterogeneity caused by the complex interactions between climate and local topography. \citet{tarca_using_2023} tracked five small Alpine glaciers individually over 2017-2021, revealing an annual area loss rate between 0.98\% and 3.4\%. Since glaciers leave behind distinct landforms (e.g. moraines) as they move across the landscape, various studies reconstructed the glacier extents during the Little Ice Age (an advanced glacier position, typically in the second half of the 19th century) and compared these to the present state, showing that glaciers decreased in size in all regions, with some glaciers that disappeared \citep{paul_glacier_2019, parkes_twentieth-century_2018}.

Glacier mapping is also an essential task due to its direct link with \gls{mb} estimation. Glacier \gls{mb}\footnote{For a more comprehensive set of definitions of glacier-related terms, see \citet{cogley_glossary_2011}.} is defined as the total sum of all the accumulation (e.g. snow, freezing rain, avalanches) and ablation (e.g. melting, calving, sublimation) across the entire glacier over a certain period (usually a year or a season). One way to estimate the \gls{mb} of the glacier is through \glspl{dem} differencing: two \glspl{dem} are co-registered, and then the elevation difference (i.e. volume change) is converted into mass change by making certain density assumptions. This technique is usually referred to as the geodetic method \citep{berthier_measuring_2023}, the resulting geodetic \glspl{mb} being an essential indicator of the glacier status. The dataset from \citet{hugonnet_accelerated_2021} provides geodetic \glspl{mb} with global coverage over the 2000-2019 period, based on \gls{dem} time-series from ASTER. When two (or more) \glspl{dem} are differenced to estimate the elevation change, it is also important to consider the potential change in the glacier area. When estimating the specific \gls{mb} (i.e. the total mass balance divided by the area), many studies assume no area change and, therefore, use a single-dated glacier outline \citep{berthier_measuring_2023}. However, this can induce significant biases, especially when the time gap between the two \glspl{dem} acquisitions is large, and the glaciers retreated significantly between the two DEMs considered. A recent study focused on five North American glaciers shows that a fixed area assumption can cause a \gls{mb} underestimation of up to 19\% \citep{florentine_how_2023}. Recent global studies on \gls{mb} estimation \citep{zemp_global_2019, hugonnet_accelerated_2021} have used regional correction factors to account for glacier area changes. Ideally, this correction should be performed independently for each glacier, but this would require inventories that temporarily match the \glspl{dem} acquisitions, which are usually unavailable at large scale. In general, in geodetic \gls{mb} studies, the glacier outlines are updated for relatively small regions, e.g. the Swiss Alps \citep{mannerfelt_halving_1931}, the European Alps \citep{sommer_rapid_2020}, a subregion from Svalbard \citep{geyman_historical_2022}, Northern \& Southern Patagonian Icefields \citep{abdel_jaber_heterogeneous_2019}. In contrast, for large regions, the outlines are typically static, e.g. for \gls{hma} \citep{brun_spatially_2017, shean_systematic_2020} or the Andes \citep{braun_constraining_2019, dussaillant_two_2019}. Similar glacier outline mismatches can also occur for in-situ \gls{mb} estimation when point-wise measurements have to be extrapolated to the entire glacier, which, therefore, also requires updating glacier outlines \citep{huss_conventional_2012}. 

Earth's large polar ice sheets in Antarctica and Greenland contain massive glaciers, storing the majority of global freshwater~\citep{siegert2005_role}. Besides most alpine glaciers, these glaciers are usually marine-terminating, meaning they terminate into the ocean, calving off icebergs. Fundamental differences such as the presence of open water and sea ice set apart mapping these glaciers from glacier mapping in lower latitudes. Therefore, calving front detection is often categorized separately from alpine glacier mapping, with the recent work on global glacier mapping by \citet{maslov_towards_2024} suggesting the use of dedicated approaches for mapping the calving fronts. Consequently, we will treat calving front detection separately in this chapter.

\subsection*{Deep Learning} Previous studies have shown that \gls{dl} can provide a significant performance improvement compared to classical approaches for a wide range of \gls{eo} tasks, including for instance, land-cover classification, vegetation parameters estimation (e.g. height,  biomass) and precipitation down-scaling or now-casting \citep{yuan_deep_2020}. Consequently, \gls{dl} was also adopted in cryospheric studies, e.g. for sea-ice concentration forecasting \citep{andersson_seasonal_2021}, glacier evolution modelling \citep{bolibar_nonlinear_2022,jouvet_deep_2022}, modelling the ice thickness of Antarctica \citep{leong_deepbedmap_2020}, estimating the mass balance of ice sheet and its various components \citep{de_roda_husman_high-resolution_2024, van_der_meer_deep_2023} and detecting blue ice in Antarctica \citep{tollenaar_where_2024}, to name a few examples. 

\begin{figure}
\centering
    \centering
    \includegraphics[trim={0.0cm 0.0cm 0.0cm 0.0cm}, clip, height=0.55\paperwidth]{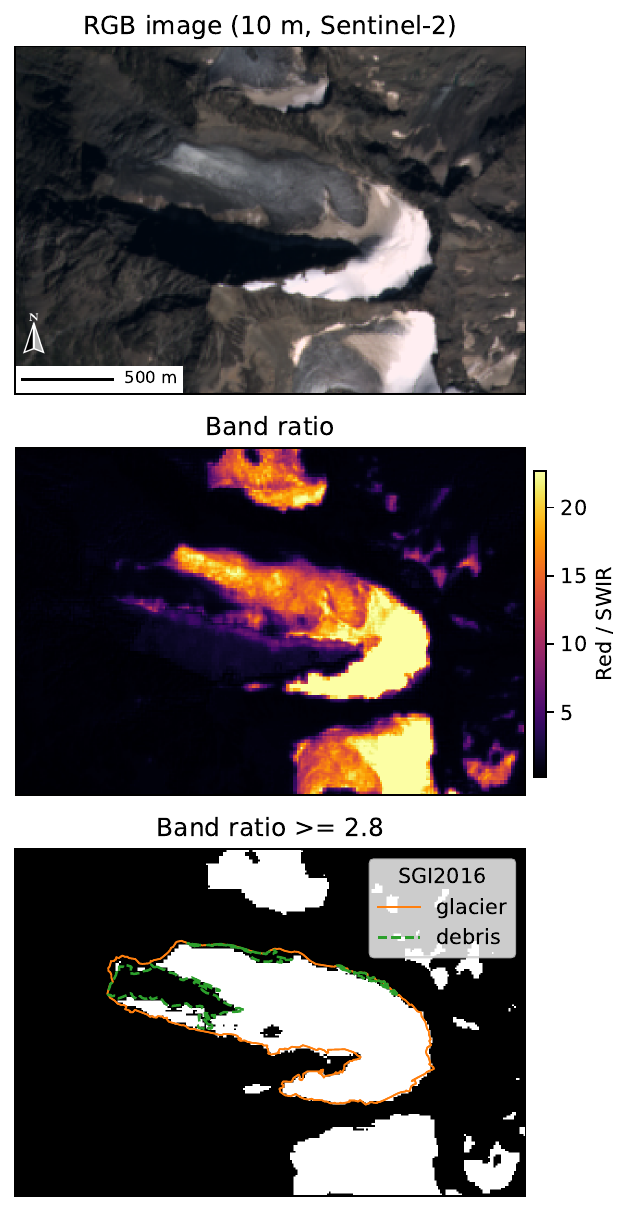}
    \captionof{figure}{\textbf{Band-ratio method.} Example of the Band-ratio method being applied to Vadret da Misaun, a glacier in Switzerland (46.42$^{\circ}$ N, 9.89$^{\circ}$ E). The upper panel uses the RGB bands from the Sentinel-2 acquisition on 26/08/2015 (courtesy of EU Copernicus program). We then apply the classical band-ratio method (central panel) using the Red and SWIR bands, with a threshold of 2.8, to delineate glacier areas (lower panel). This threshold was manually chosen to extract shadowed pixels. When comparing to the glacier \& debris outlines from the \gls{sgi} \citep{linsbauer_new_2021}, we notice that the clean-ice and snow are accurately extracted, but this approach does not capture the debris-covered parts.}
    \label{fig:s2-band-ratio}
\end{figure}

A particular area of cryospheric sciences where \gls{dl} is rapidly evolving is the domain of glacier mapping. First, given the large number of glaciers, fully automating the task of mapping glacier outlines is of high interest. Historically, semi-automatic methods were used for glacier mapping, which usually consist of thresholding band ratios, e.g. Red / \gls{swir}, as exemplified in \Cref{fig:s2-band-ratio}, or the \gls{ndsi}. This thresholding step is usually followed by manual corrections, especially for debris-covered glacier parts or those subject to shadowing \citep{paul_new_2002, paul_glacier_2020}. Similarly, rough estimates of calving fronts can be obtained from thresholding reflectance or backscatter values, as the ocean will appear much darker than ice and snow in imagery. However, such approaches are easily confounded by the presence of sea ice \citep{liu2004_complete}. A particular difficulty of these semi-automatic methods relates to the fact that, in many cases, thresholds need to be calibrated to each different scene, e.g. to account for varying atmospheric conditions or to capture shadowing effects. From this perspective, \gls{dl} has the added value of potentially exploiting local features and automatically finding the appropriate threshold. Second, even though \gls{dl} models have associated errors, one can assume that these errors are systematic in space and time (assuming an in-distribution testing scenario). In contrast, the quality of the manual glacier outlines can significantly vary depending on the individuals ('experts') performing the glacier labelling \citep{paul_accuracy_2013}. To reduce these errors related to manual labelling in newer inventories, a preparation phase is usually implemented to ensure a consistent and homogeneous quality among the human annotators \citep{linsbauer_new_2021}.  Third, \gls{dl} can learn from multiple data sources simultaneously, which is particularly helpful for more complex classification cases. Such an example is classifying debris-covered glaciers \citep{xie_glaciernet_2020} that are visually very difficult to distinguish from their (non-glaciated) surroundings, which classical band-ratio methods can only (partly) circumvent through time-consuming and error-prone manual corrections \citep{paul_glacier_2020}.

Despite the enormous potential of \gls{dl} to be applied in the field of glacier mapping, some substantial challenges exist depending on the scientific question of interest. First, despite continuous improvements, the glacier inventories needed for training \gls{dl} architectures suffer from considerable uncertainties. Although \gls{dl} can deal with "noisy" labels to a certain extent and still learn the underlying patterns \citep{arpit_closer_2017, zlateski_importance_2018}, these uncertainties certainly affect the performance metrics used when evaluating the quality of the predictions, making it more challenging to compare different methods. Additionally, the uncertainties in the inventories also vary from one region to another, thus potentially introducing biases when training global models. Moreover, in many cases, it is difficult (and often even entirely impossible) to ensure a perfect temporal match between the (optical) input data and the glacier inventory, which adds to the uncertainties. Lastly, glacier mapping remains a challenging task even for experts, especially for debris-covered glaciers where the interpretation can be subjective, sometimes leading to errors in the order of 10\%-20\% for small glaciers \citep{paul_glacier_2020}. These limitations should be accounted for when developing adequate \gls{dl} frameworks for glacier mapping, and associated uncertainties in \gls{dl} model predictions should always be quantified \citep{maslov_towards_2024}. 

\FloatBarrier
\section{Data Modalities}
\label{sec:data-modalities}

This section summarises the data modalities exploited by various \gls{dl} models from the literature, surveyed in \Cref{sec:literature-overview}, or used by experts when building glacier inventories.


\subsection*{Optical (multi-spectral) imagery} Optical data\footnote{Note that we here mainly refer to medium resolution imagery (10 to 30 m \gls{gsd})} is by far the most commonly used type of observation for glacier mapping, primarily because in many circumstances (e.g. for clean-ice glaciers), optical data can already support glacier delineation on its own, without the need of additional data sources. The importance of optical data relative to some other modalities (i.e. \gls{sar} backscatter intensities, \gls{insar} coherence, \gls{dem}, thermal imaging) was empirically evaluated using ablation studies in the works of \citet{peng_automated_2023} and \citet{maslov_towards_2024}. Optical data alone helps building very accurate glacier outlines, especially for debris-free glaciers where the classical band-ratio method or \gls{ndsi} thresholding yields robust results. The band-ratio method (e.g. Red / \gls{swir}) can separate the very low spectral reflectance of ice and snow in the shortwave infrared versus the high reflectance in the visible spectrum \citep{paul_accuracy_2013, paul_glaciers_2015}.  From a \gls{dl} perspective, optical data offers many advantages: i) numerous data sources available, including many with open-access policies, e.g. Sentinel-2 or Landsat-9 (or older), ii) most of the \gls{dl} architectures were designed for optical data (usually RGB), with many pre-trained models publicly available, and iii) it is relatively easy to visualize and understand this data, even for non-experts, either by extracting the RGB bands or by using false colour images. However, optical data comes with some disadvantages, i.e. i) many glaciers are often covered by clouds, limiting the amount of data, ii) optical data over glaciers often has decreased visibility due to shadows (either caused by clouds or surrounding topography) or suffers from the absence of sunlight (e.g. the case of glaciers in the polar areas during the Polar Night), iii) optical data is relatively sensitive to atmospheric interference, iv) optical data cannot, in general, distinguish between debris-covered glacier segments and the off-glacier surrounding topography, and v) typically data from the end of summer is needed to reduce the presence of seasonal/perennial off-glacier snow. 

\begin{figure}
\centering
    \centering
    \includegraphics[trim={0.0cm 0.0cm 0.0cm 0.0cm}, clip, width=0.7\textwidth]{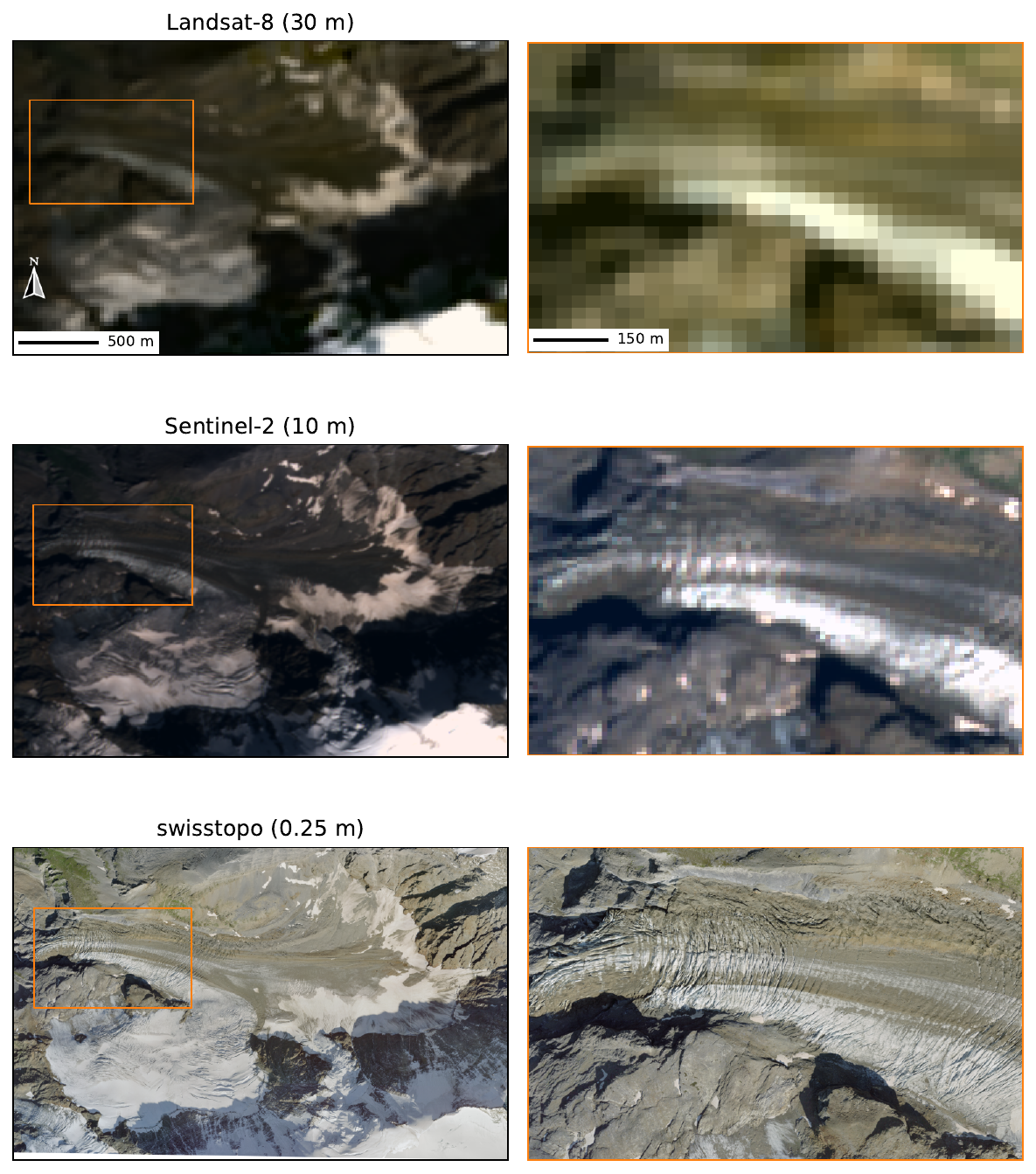}
    \captionof{figure}{\textbf{Spatial resolution comparison.} The effect of spatial resolution on visual products is here illustrated for Rottalgletscher, a glacier in Switzerland (46.52$^{\circ}$ N, 7.95$^{\circ}$ E). We use the RGB bands from the Landsat-8 acquisition on 22/08/2018 (image courtesy of the U.S. Geological Survey), from Sentinel-2 on 20/08/2018 (courtesy of EU Copernicus program) and the \gls{vhr} aerial image from \citet{swisstopo_swissimage_2024} (flight year = 2018). As resolution increases, an increasing level of detail can be observed, particularly pronounced for the debris-covered and shadowed glacier parts. The right panels provide a zoomed-in view of the glacier tongue, illustrating, among others, how the crevasses, a distinct feature of glaciers, become visible with increasing spatial resolution. }
    \label{fig:res-comparison}
\end{figure}

However, some of these disadvantages can be compensated by using \gls{vhr} optical data, if available. In \Cref{fig:res-comparison}, we show optical data over a single glacier at three different spatial resolutions, i.e. 30 m, 10 m and a \gls{vhr} one, 25 cm. From this, we can, for instance, observe that the debris-covered segments become distinguishable in the \gls{vhr} image due to the presence of crevasses, which are hardly visible in the other (lower resolution) products. Additionally, we can notice in the \gls{vhr} data that the segments under shadow are still relatively visible compared to the other, lower-resolution sensors. \citet{paul_accuracy_2013} compare the glacier outlines from multiple analysts on two resolutions, i.e. 30 m and 1 m, and suggest that the interpretation of debris-covered segments is mainly independent of the resolution.  It is, however, hard to conclude from such studies whether \gls{dl} models would perform differently, and we did not find any study that analyzes the role of spatial resolution from this perspective. None of the existing \gls{dl} methods discussed in \Cref{sec:literature-overview} has utilized \gls{vhr} data, suggesting the limited availability and high costs of such high-resolution data. However, as the volume of (open) data continues to increase over time, we anticipate a growing number of studies exploring these avenues in the future.

\subsection*{Synthetic Aperture Radar (SAR)} 
\begin{figure}
\centering
    \centering 
    \includegraphics[trim={0.0cm 0.0cm 0.0cm 0.0cm}, clip, height=0.45\paperwidth]{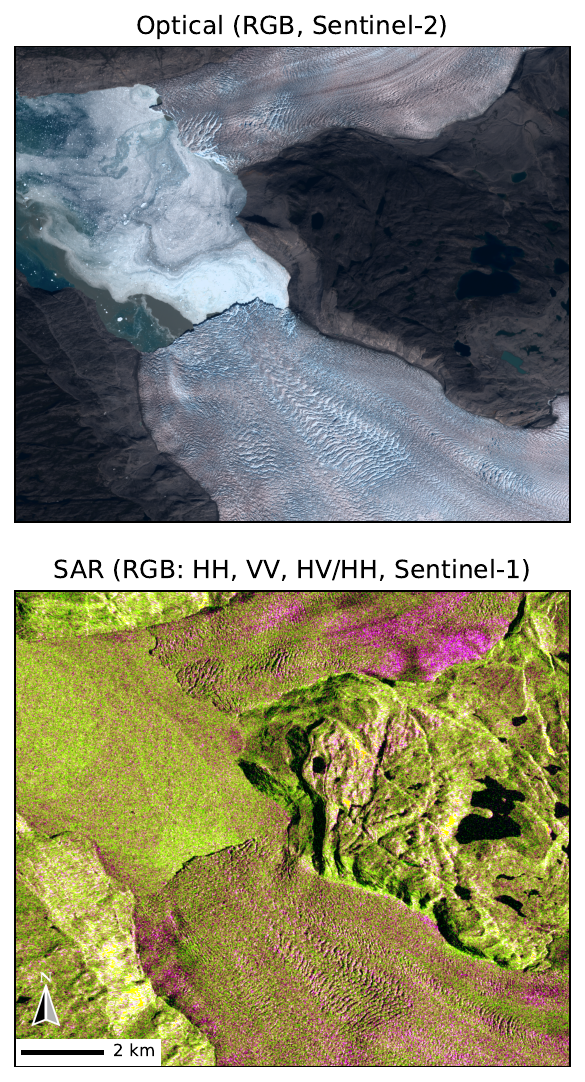}
    \captionof{figure}{\textbf{Calving fronts - optical and \gls{sar}.} Two calving fronts that belong to the Kangiata Nunaata Sermia and Akullersuup Sermia glaciers, in Greenland (64.33$^{\circ}$ N, -49.65$^{\circ}$ E), are observed using optical data (upper panel, RGB bands from the Sentinel-2 acquisition on 24/07/2023), and \gls{sar} data (lower panel, false-color composite extracted from the Level-1 Ground Range Detected of the Sentinel-1 acquisition on 06/07/2023. Courtesy of EU Copernicus program.} 
    \label{fig:cf}
\end{figure}

Although not widely adopted yet, various types of \gls{sar} datasets have been explored in the glacier mapping literature. Compared to optical sensors, they offer various advantages: i) \gls{sar} uses longer wavelengths that have the potential to "see" through clouds, an important benefit for mountain regions; ii) being an active remote sensing method, it has day/night capability; iii) it can penetrate the snow (to some extent), which can be beneficial when transient snow obscures the boundaries of glaciers. \gls{sar}, however, also has some disadvantages, primarily due to the side-looking geometry, resulting in radar shadowing and foreshortening/layover, which are especially problematic in (steep) mountain regions. Additionally, for \gls{dl} practitioners, \gls{sar} is more difficult to interpret than optical data and requires more intensive preprocessing. Lastly, \gls{sar} speckle patterns may cause additional challenges to the traditional \gls{dl} architectures that are usually designed for optical (RGB) data \citep{zhu_deep_2021}. An example of a \gls{sar} intensity image with its optical correspondent is displayed in \Cref{fig:cf}.

While optical data is often sufficient for debris-free glaciers to distinguish glacier outlines under cloud- and snow-free conditions, \gls{sar} can play an important role for the debris-covered segments and, by extension, for rock glaciers. First, \gls{sar} backscatter intensities correlate with the surface roughness, which can help distinguish the debris parts from the surrounding terrain. Second, by combining two (or more) \gls{sar} acquisitions, \gls{insar} can reveal glacier motion or deformations, an indicator of glacier segments covered by debris or active rock glaciers. Previous studies have employed either \gls{insar} displacement maps using the (unwrapped) interferogram or  \gls{insar} coherence \citep{robson_automated_2020, maslov_towards_2024}.

\subsection*{Digital Elevation Model} 
\begin{figure}
\centering
    \centering
    \includegraphics[trim={0.0cm 0.0cm 0.0cm 0.0cm}, clip, width=0.4\textwidth]{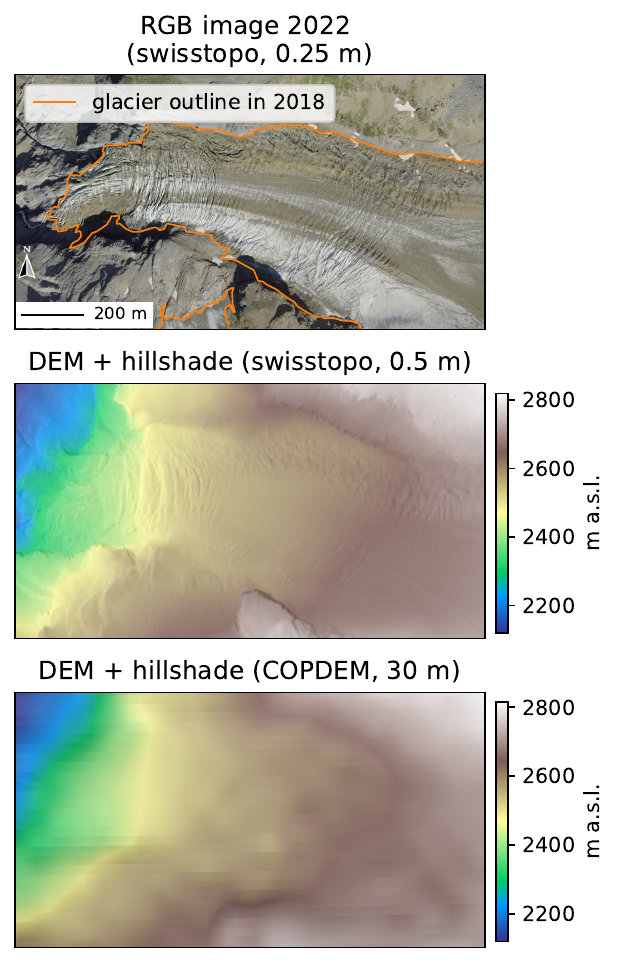}
    \captionof{figure}{\textbf{\glspl{dem} at two different spatial resolutions}. The figure displays the tongue of the Rottalgletscher glacier in Switzerland (46.52$^{\circ}$ N, 7.94$^{\circ}$ E) (see also \Cref{fig:res-comparison}). The \glspl{dem} are extracted from the swissALTI$^{\text{3D}}$ \gls{dem} \citep{swisstopo_swissalti3d_2024} and the Copernicus GLO-30 \gls{dem}, and are here displayed with a superimposed shaded relief. While both \glspl{dem} capture the shape of the valley the glacier flows in, only the \gls{dem} with a sub-meter resolution captures smaller scale features such as crevasses.}.
    \label{fig:dem}
\end{figure}

The surface elevation is usually provided as an additional input for glacier mapping efforts since it helps the models to extract further information related to topography, e.g. the glacier flow direction, which can be a valuable source of information, especially for calving glaciers. Additionally, using the elevation information, the \gls{dl} models can "understand" where the glacier terminus is, which can be covered by debris as opposed to the accumulation areas. However, obtaining a large-scale \gls{dem} from satellite data requires specialized sensors and processing techniques, with additional challenges in mountain regions. For instance, two standard techniques are stereo photogrammetry and \gls{insar}. If the former is affected by cloud and snow coverage or topographic shadowing \citep{hugonnet_accelerated_2021}, the latter is challenged by the steep terrain and can yield significant ice-penetration biases \citep{dehecq_elevation_2016, berthier_measuring_2023}. Studies that rely on \gls{dem} for glacier mapping often need to account for a mismatch in timing between the \gls{dem} and the considered visual imagery. Despite this limitation, \gls{dem} information can still provide helpful information about the glacier topography and its surroundings. A few standard and openly available \gls{dem} choices are Copernicus GLO-30 DEM (Cop30DEM), Shuttle Radar Topography Mission (SRTM) DEM or its improved version - NASADEM, and ALOS World 3D - 30m (AW3D30). \Cref{fig:dem} displays a \gls{dem} at two different spatial resolutions, i.e. 30m and 0.5m, for the tongue of the glacier from \Cref{fig:res-comparison}. While both \glspl{dem} roughly capture the valley in which the glacier flows, crevasses become partially visible in the \gls{vhr}, which can help to better identify the debris-covered parts. This comparison again suggests that spatial resolution could affect the models' performance.

\begin{figure}
\centering
    \centering
    \includegraphics[trim={0.0cm 0.0cm 0.0cm 0.0cm}, clip, height=0.6\textheight]{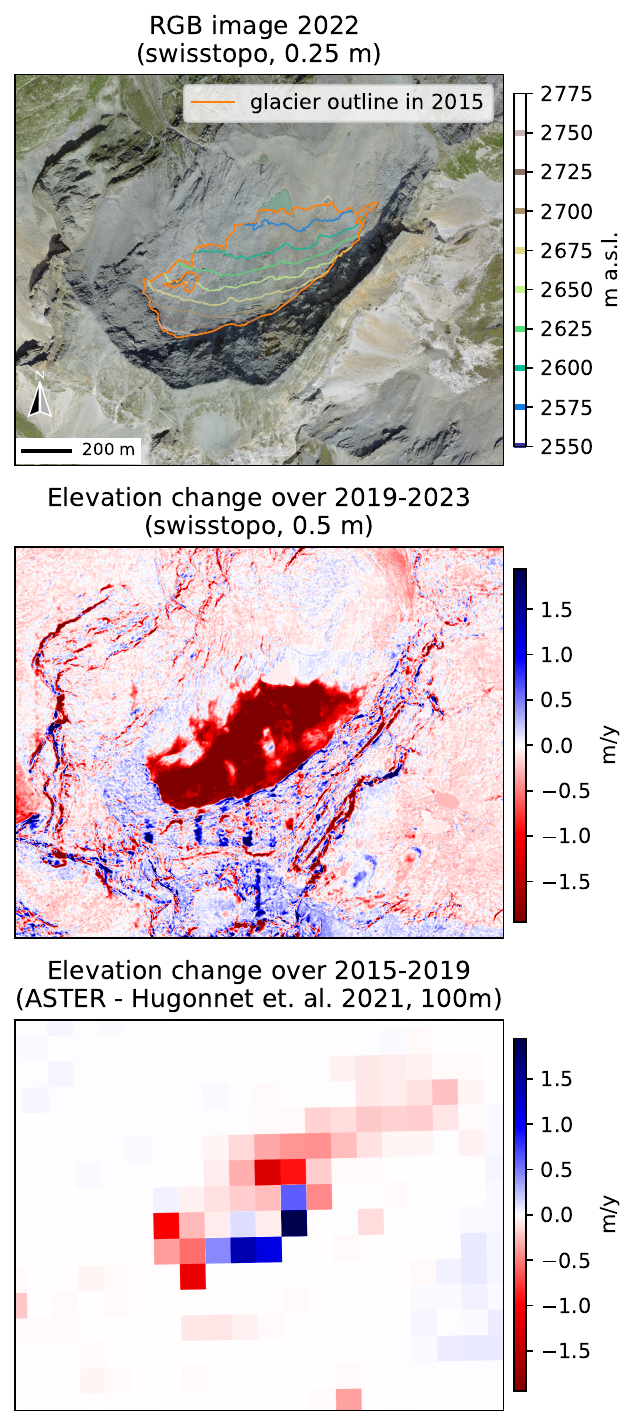}
    \captionof{figure}{\textbf{Differencing of \glspl{dem}}. Here, we illustrate the role of resolution in \gls{dem} differencing for an entirely debris-covered glacier, Glatscher da Sut Fuina, in Switzerland (46.535$^{\circ}$ N, 9.473$^{\circ}$ E). While both \gls{dem} differences allow identifying the location of the glacier, the \gls{vhr} version (central panel), based on two swissALTI$^{\text{3D}}$ \glspl{dem} \citep{swisstopo_swissalti3d_2024}, is significantly more accurate. Note that for this \gls{vhr} \gls{dem} differencing, the two \glspl{dem} were not co-registered before differentiation, which would allow for some artefacts to be removed. The \gls{dem} differencing map based on the lower-resolution \glspl{dem} \citep{hugonnet_accelerated_2021} (lower panel) is only able to roughly capture the location of the glacier, with some potential outliers (e.g., the significantly positive pixels). This figure illustrates that when using \gls{dem} differencing to detect glaciers, the role of the spatial resolution becomes important for relatively small glaciers.}
    \label{fig:dhdt}
\end{figure}

When at least a pair of \glspl{dem} is available, it is possible to derive a surface elevation change map. For instance, a negative elevation change rate was observed over the last two decades (2000-2019) for almost all the glaciers in the world \citep{hugonnet_accelerated_2021}. As a result, \glspl{dem} differences can help capture the glacier extent by contrasting it to the surrounding topography where typically no change is expected. This information can supplement image classification approaches, especially for debris-covered glaciers, as these are difficult to classify using optical data alone. However, the availability of global products is somewhat limited: the only global product that is based on the same data source (i.e. ASTER) from \citet{hugonnet_accelerated_2021} comes at 100 m \gls{gsd}. Moreover, for short timescales, high-precision \glspl{dem} are needed, both vertically and horizontally, to distinguish between potentially small glacier surface changes and off-glacier changes due to noise. In \Cref{fig:dhdt}, we highlight how a \gls{vhr} elevation change map helps identify glaciers that are entirely covered by debris. 

Ideally, we want to track glacier area change over time, e.g. at an annual to sub-decadal timescale, to capture how glaciers respond to changes in climatic conditions. Therefore, temporal resolution also plays a vital role in all the aforementioned data sources. One difficulty arises from the regions that undergo deglaciation early in the period spanned by two acquisitions: for these regions, which are of particular interest when tracking glacier areas over time, without a time-series, it is not feasible to identify the segments that become deglaciated early-on during the covered period, thereby hindering the analysis on the involved response times.

\paragraph*{}
In summary, we have discussed various data sources for glacier mapping, each offering its pros and cons. This comparison emphasizes the potential of integrating multi-sensor data into \gls{dl} models. As we explain in the next section's literature overview, most studies utilize at least two data sources.

\FloatBarrier
\section{Literature Overview}
\label{sec:literature-overview}

This section briefly reviews some of the key studies that employed \gls{dl} models for automatic glacier delineation. Note that this is not an exhaustive review of all existing methods but rather a short overview of some of the most relevant and innovative works, emphasizing the particular challenges that have been addressed and shedding light on the obstacles that will require further research.  Additionally, we highlight the various data sources used in each work to illustrate the importance of relying on multi-sensor approaches when mapping glaciers through \gls{dl}. Given the significant differences in input data sources, labels, and/or considered regions, the evaluation scores mentioned throughout this section should not be overinterpreted or used to compare the performance of the various methods. 

The section is structured as follows. First, we describe the works on glacier extent mapping in \Cref{sec:glacier-extent-mapping}, with three sub-categories: i) standard methods, i.e. those that focus in general on glacier extent mapping with the primary goal of automatizing the process, ii) studies that perform glacier mapping on multiple acquisitions to quantify temporal glacier area changes and, iii) studies that map the extent of rock glaciers, which we treat as a separate category given their significant differences compared to typical glaciers, thereby usually requiring more specialized methodologies. The various studies and corresponding methods are then summarized in \Cref{tab:glacier-extent-mapping-standard,tab:glacier-extent-mapping-area-change,tab:glacier-extent-mapping-rock-glaciers}, respectively. The second part covers the works on calving front detection, which are then summarized in \Cref{tab:cf-detection}. Note that most of the paragraphs cover a single study, with some exceptions where follow-ups are also included.

\subsection{Glacier Extent Mapping}
\label{sec:glacier-extent-mapping}

\subsubsection{Glacier Extent Mapping - Standard Methods}
\label{sec:glacier-extent-mapping-standard}

The methods included in the following paragraphs treat the glacier extent mapping problem as a single-image segmentation task and propose various modifications to existing \gls{dl} architectures, usually with the goal of improving the performance for the debris-covered parts of the glacier which remain hard to detect. One common characteristic of these methods is their data fusion capability: as highlighted in \Cref{tab:glacier-extent-mapping-standard}, most works use at least two data sources, with optical data being the main one.

\paragraph*{GlacierNet \citep{xie_glaciernet_2020}}
The first work that employs a \gls{dl} architecture is GlacierNet \citep{xie_glaciernet_2020}. Whereas previous studies that used \gls{ml} relied on classical approaches, e.g. support vector machine (SVM), random forest (RF) or shallow networks like multi-layer perceptrons (MLPs) \citep{zhang_glacier_2019, khan_machine-learning_2020}, GlacierNet proposes a \gls{cnn} architecture built upon the fully convolutional model SegNet \citep{badrinarayanan_segnet_2017}. The main purpose of the work was to address the challenge of detecting debris-covered glaciers. It was trained using all eleven Landsat 8 bands and the AW3D30 \gls{dem}, from which additional features were derived, i.e. slope angle, slope-azimuth divergence index (SADI), profile curvature, tangential curvature and unsphericity curvature. The study focuses on two sub-regions from \gls{hma} with different glacier conditions and properties: Nepal Himalaya and the central Karakoram in Pakistan. The glacier boundaries used for training were obtained from \gls{glims} \citep{raup_glims_2007} and further modified i) by improving the termini delineations, which can be at a different position due to the time mismatch between the imagery and the inventory, and ii) by removing the snow-covered accumulation zones, as it is indistinguishable from the surrounding snow-covered terrain. Once the model is trained and inferences are made, a post-processing step is applied to improve the predictions by i) connected-region size thresholding, ii) gap-filling, iii) improving the predictions over lake-contact termini based on the \gls{ndwi}. Using only the imagery and the \gls{dem} as inputs, the method achieves 80.9\% \gls{iou} and 89.5\% F1 on the testing regions, which further increase to 84.1\% and 91.4\%, respectively, when including all the \gls{dem}-derived features.  
 
 In a different study focused on the central Karakoram region, \citet{xie_evaluating_2021} compare GlacierNet \citep{xie_glaciernet_2020}  against five different \gls{cnn}-based segmentation models. Instead of \gls{glims}, the authors used for the ablation areas the more-accurate contours from the \gls{gamdam} dataset \citep{nuimura_gamdam_2015}. The chosen baseline methods include three versions of U-Net \citep{navab_u-net_2015}: Mobile-UNet \citep{jing_mobile-unet_2022} i.e. U-Net with a MobileNetV2  backbone \citep{sandler_mobilenetv2_2018},  Res-UNet i.e. U-Net with a ResNet34 backbone \citep{he_deep_2016}, and R2U-Net \citep{alom_recurrent_2018} i.e. U-Net with recurrent convolutional layers. The last two are FCDenseNet \citep{jegou_one_2017} and DeepLabv3+ \citep{chen_encoder-decoder_2018} with an Xception backbone \citep{chollet_xception_2017}. They found that DeepLabv3+ performs the best, with an 86.2\% \gls{iou}. However, GlacierNet gives the second-highest score, only 0.2\% lower, but at a smaller computational cost.

\paragraph*{GlacierNet2 \citep{xie_glaciernet2_2022}}
In a follow-up study, \citet{xie_glaciernet2_2022} proposed an improvement of GlacierNet by i) using multiple models to improve the predictions over the glacier termini, ii) including also the snow-covered accumulation zones, which were previously discarded. GlacierNet2 can be considered as a two-member ensemble model as it combines the previous GlacierNet with the predictions from a DeepLabv3+ model \citep{chen_encoder-decoder_2018} with an Xception backbone \citep{chollet_xception_2017}. Each sub-model is trained independently, then their weights are frozen, and lastly, a final 1x1 convolutional layer is trained to fuse their predictions. However, the predictions from GlacierNet alone are still kept and post-processed in parallel with those from the fused GlacierNet-DeepLabv3+ model. Adding to the post-processed steps proposed in the previous work (i.e. gap-filling and region-size thresholding), the authors implement an additional step where the final predictions of the two sub-components are compared at the termini and disagreements are addressed through a k-Nearest Neighbor (KNN) classifier. Also, a particular post-processing pipeline is applied to the accumulation areas for the snow-covered pixels detected using the \gls{ndsi}. An ablation study shows that the final model reaches an 88.4\% \gls{iou} and a 93.8\% F1 score, improving by 1-2\% the baselines, i.e. GlacierNet, DeepLabv3+ and their combination, when evaluated on ablation-zone mapping.

\paragraph*{\citet{tian_mapping_2022}} An improved U-Net \citep{navab_u-net_2015} architecture is used by \citet{tian_mapping_2022} to segment glaciers in the Pamir Plateau. The original architecture is improved by incorporating the channel attention module from  \citet{frangi_concurrent_2018}, the so-called channel squeeze and excitation block. On top of this, the authors also used a conditional-random field (CRF) method as post-processing to refine the results. For training, they used optical data from Landsat-8 and the SRTM \gls{dem} with the labels based on \gls{glims} \citep{raup_glims_2007}. These were manually corrected to account for a temporal gap between the imagery and the labels during which glacier changes occurred. The model achieved an F1 score of 89.5\% after adding the attention mechanism, further improved to 89.8\% with the CRF-based refinement, performing better than the original U-Net, which obtained 88.9\%. They also tested the GlacierNet model \citep{xie_glaciernet_2020}, which obtained an F1 score of only 84.9\%.

\paragraph*{\citet{chu_glacier_2022}} Similar to the previous study of \citet{tian_mapping_2022}, in this work, \citet{chu_glacier_2022} also incorporate an attention mechanism, i.e. convolutional block attention module (CBAM) \citep{woo_cbam_2018}. They build upon the DeepLabv3+ architecture \citep{chen_encoder-decoder_2018} with a ResNet-34 backbone \citep{he_deep_2016}. Additional improvements were obtained using test-time augmentation and depth-wise separable convolution at the end of the segmentation \citep{chollet_xception_2017}. They used high-resolution (8m) multi-spectral data (R, G, B, \gls{nir}) from the Gaofen-6 satellite, which also provides a panchromatic band with 2m resolution. For training, they manually delineated glaciers from the Tanggula, Kunlun and the Qilian Mountains. Finally, they compared their results based on the model's predictions with existing regional inventories, showing a more accurate extraction of the debris-free glaciers. Their model achieves an F1 score of 98.54\%. They compared to multiple baselines, the best performing one being the original DeepLabv3+ model, which achieves already an F1 score of 98.3\%, followed by U-Net \citep{navab_u-net_2015} with a ResNet-18 backbone \citep{he_deep_2016} which obtains an F1 score of 97.8\%. 

\paragraph*{\citet{peng_automated_2023}} The transformer-based architecture builds upon Swin-Unet \citep{karlinsky_swin-unet_2023}, with a series of improvements from various other works. The decoder was coupled with locally-grouped self-attention and global sub-sampled attention modules from Twins-SVT \citep{chu_twins_2021}, together with the conditional position encoding from \citet{chu_conditional_2022}. The decoder uses Local-Global CNN Blocks, ending with a Feature Refinement Head as the segmentation head, both from Unet-Former \citep{wang_unetformer_2022}. The study focuses on the Qilian Mountains, using the inventory from \citet{li_glacier_2020}. As input data, the following were used: five optical bands (Sentinel-2), i.e. R, G, B, \gls{nir} and \gls{swir} (B11), from which three indices were obtained i.e. \gls{ndvi}, \gls{ndwi} and \gls{ndsi}; two \gls{sar} backscatter intensity images (VV and VH polarizations), using the Level-1 Ground Range Detected from Sentinel-1; and a \gls{dem} using the 8m  High Mountain Asia Digital Elevation Model (HMADEM) \citep{shean_high_2017} as the main source, complemented with the SRTM one for the regions not covered by the former. The final model achieves an F1 score of 84.3\%, followed by the Swin transformer \citep{liu_swin_2021} with 82.9\% and the model from the previous study of \citet{chu_glacier_2022}, based on DeepLabv3+ \citep{chen_encoder-decoder_2018}, with 82.2\%. A detailed ablation study on various input features groups shows that optical bands are the most important but also highlights the benefit of combining multi-source datasets.

\paragraph*{\citet{thomas_integrated_2023}} This work focuses on mapping debris-covered glaciers from three \gls{hma} sub-regions, i.e. Hunza (Karakoram), Manaslu (Central Himalayas) and the Khumbu (Central Himalayas). The methodology is similar to the one from a previous study (discussed in \Cref{sec:glacier-extent-mapping-rock-glaciers}), focused on mapping rock glaciers \citep{robson_automated_2020}. It consists of a five-layer \gls{cnn} with a post-processing stage using \gls{obia}. The model's output is finer compared to previous studies, as it distinguishes among seven classes, i.e. supraglacial debris, clean ice, snow cover, lakes, vegetation, shadows, and non-glacial material. As input, the model uses a wide range of features (21 in total):
\begin{itemize}
    \item ten optical bands (Sentinel-2) from which three indices were obtained i.e. \gls{ndvi}, \gls{ndwi} and \gls{ndsi}
    \item \gls{sar} coherence (Sentinel-1)
    \item a thermal band (Landsat-8)
    \item a \gls{dem} (AW3D30) with five derived features, i.e. slope angle, profile curvature, planform curvature, aspect, and shaded relief
\end{itemize}
The model is trained using the \gls{gamdam} inventory \citep{nuimura_gamdam_2015}, and the analysis of the results shows that the integration of \gls{obia} as a post-processing step helps reduce the model's errors.
Additionally, for one of the regions (Manaslu), the authors apply their methodology to declassified panchromatic imagery from the Corona KH-4B satellite. Although the results show the limitations caused by the absence of the multi-spectral dimension, the approach illustrates the potential towards building historical inventories for \gls{hma}. 

\paragraph*{GlaViTU \citep{maslov_glavitu_2023}}
Following the advances in Computer Vision research, an architecture based on a VisionTransformer (ViT) \citep{dosovitskiy_image_2020} is used. The authors propose a hybrid \gls{cnn}-transformer model, called Glacier-VisionTransformer-U-Net or GlaViTU, by combining  SEgmentation TRansformer (SETR) \citep{zheng_rethinking_2021} with a ResNet-backbone U-Net \citep{navab_u-net_2015, he_deep_2016}. The method is compared to three baselines. First, TransUNet \citep{chen_transunet_2021}, i.e. a different type of U-Net that uses a hybrid  ResNet-50 \citep{he_deep_2016} + ViT \citep{dosovitskiy_image_2020} encoder. The second, ResU-Net, is the U-Net with a classical ResNet backbone, also used in the previously mentioned study of \citet{xie_evaluating_2021}. The third baseline is a SETR-B/16 model
with progressive upsampling decoder \citep{zheng_rethinking_2021}. For a fair comparison, all the baselines are modified by adding a data-fusion \gls{cnn}-based head as in the proposed architecture. This block was proposed to better fuse the three input data modalities, i.e. multi-spectral optical data from Landsat 7, 8 and Sentinel-2, $\sigma_0$-calibrated amplitude images from Envisat and Sentinel-1, and \glspl{dem} (Cop30DEM and AW3D30). The dataset is much larger compared to previous studies, and it covers six regions worldwide: the European Alps, High-Mountain Asia, Indonesia, New Zealand, the Southern Andes and Scandinavia.  On average, over these six regions, GlaViTU achieves an 87.5\% \gls{iou}, followed by TransU-Net and ResU-Net with 86.3\% and 85.2\%, respectively.

In a more recent study, \citet{maslov_towards_2024} extend the previous dataset towards a global one. Compared to the previous studies, the dataset is much larger, covering most of the glacier regions outside the two ice sheets. With around 19,000 glaciers, it covers ca. 7\% of the total glaciated area. The training labels are mainly based on GLIMS \citep{raup_glims_2007} and RGI \citep{pfeffer2014randolph}, but for 8 out of the 23 sub-regions, local inventories are used. In addition to the data sources used in their previous work, i.e. optical, \gls{sar} and \gls{dem}, the authors also investigate whether adding thermal data when available from Landsat-8 helps but did not find it effective. Moreover, in this work, \gls{insar} coherence images are also considered (when available) instead of the amplitude images, improving performance. The original GlaViTU model is further improved by updating the data fusion block, which is now equipped with feature weighting and includes squeeze-and-excitation blocks. The DeepLabv3+ model \citep{chen_encoder-decoder_2018}, with a ResNeSt-10 backbone \citep{zhang_resnest_2022}, is used as a baseline, after adding the same data fusion block as for GlaViTU. The average \gls{iou} over all the regions is 89.4\%, almost 2\% higher than the baseline, which obtains 87.7\%. Instead of training a single global model, \citet{maslov_towards_2024} also investigate four additional training strategies, i.e. regional training, finetuning and location encoding (by using either the region or the coordinate), showing that on average, the regional and the fine-tuned models perform the best, with almost 1\% better than the global model. Lastly, it was investigated for the first time whether uncertainty estimation techniques, namely Monte Carlo dropout \citep{gal_dropout_2016} and temperature scaling \citep{guo_calibration_2017}, can be used to get further insights into the predictions. They concluded that temperature scaling alone could already help to extract confidence intervals for the predictions and qualitatively found that the model is more uncertain on the debris-covered segments or those under the shadow, thus illustrating the practicality of the uncertainty estimates. So far, this study represents the most comprehensive dataset for glacier mapping using \gls{dl} which can facilitate further methodological developments by directly using the processed data and the same validation procedure.\\

\begin{table}[!ht]
    \centering
        \begin{NiceTabular}{|m{0.12\textwidth}|m{0.2\textwidth}|m{0.25\textwidth}|m{0.15\textwidth}|m{0.07\textwidth}|}
    \toprule
    \RowStyle[bold]{}
    
    \Block{1-1}{Publication (model name)} &
    \Block{1-1}{model architecture\tablefootnote{Note that sometimes changes to the original architecture are made, see \Cref{sec:glacier-extent-mapping-standard}.}} &
    \Block{1-1}{data modality\tablefootnote{Often features are further derived e.g. \gls{ndvi} from optical or slope from \gls{dem}.} (source)} &
    \Block{1-1}{\acrshort{roi}\tablefootnote{We only indicate the main \glspl{roi} from where the training data was extracted. Still, the coverage can vary significantly, e.g. from a few image scenes to almost complete coverage.} } &
    \Block{1-1}{code/ data\tablefootnote{This refers to the processed training data, not the raw one. The latter is usually openly available from the original source.}/ output\tablefootnote{In case the study has an output product (e.g. an inventory).}}  \\
    
    \midrule

    \Block{1-1}{\citet{xie_glaciernet_2020} (GlacierNet)} &
    \Block{1-1}{SegNet \citep{badrinarayanan_segnet_2017}} &
    \Block[l]{2-1}{
        \begin{itemize}[leftmargin=0.25cm, itemsep=-0.4ex]
            \item optical (Landsat 8)
            \item \gls{dem} (AW3D30)
        \end{itemize}
        } & 
    \Block{1-1}{\gls{hma}: Nepal Himalaya, \\ central Karakoram} & 
    \Block[c]{1-1}{-} \\
    
    \cmidrule{1-2}\cmidrule{4-5}

    \Block{1-1}{\citet{xie_glaciernet2_2022} (GlacierNet2)} &
    \Block{1-1}{SegNet \& DeepLabv3+ \citep{badrinarayanan_segnet_2017, chen_encoder-decoder_2018}} & 
     & 
    \Block{1-1}{\gls{hma}: central Karakoram} &
    \Block[c]{1-1}{-} \\

    \midrule

    \Block{1-1}{\citet{tian_mapping_2022}}  & 
    \Block{1-1}{U-Net \citep{navab_u-net_2015} with attention} &
    \Block{1-1}{
        \begin{itemize}[leftmargin=0.25cm, itemsep=-0.4ex]
             \item optical (Landsat 8)
            \item \gls{dem} (SRTM)
        \end{itemize}
    } & 
    \Block[c]{1-1}{\gls{hma}: Pamir} & 
    \Block[c]{1-1}{-} \\

    \midrule

    \Block{1-1}{\citet{chu_glacier_2022}}  & 
    \Block{1-1}{DeepLabv3+ \citep{chen_encoder-decoder_2018} with attention} &
    \Block{1-1}{
        \begin{itemize}[leftmargin=0.25cm, itemsep=-0.4ex]
            \item optical (Gaofen-6)
        \end{itemize}
    } & 
    \Block[c]{1-1}{\gls{hma}: Tanggula, Kunlun, Qilian} &
    \Block{1-1}{\href{https://github.com/yiyou101/Attention-DeepLab-V3plus}{C}/\href{https://pan.baidu.com/s/1P0FFkq3zrIbYfDVTLC_soA?pwd=ctsa}{D}/-}
    \\

    \midrule

    \Block{1-1}{\citet{peng_automated_2023}}  & 
    \Block{1-1}{based on Swin-Unet \citep{karlinsky_swin-unet_2023}, with various improvements} &
    \Block{1-1}{
        \begin{itemize}[leftmargin=0.25cm, itemsep=-0.4ex]
            \item optical (Sentinel-2)
            \item \gls{sar} (Sentinel-1)
            \item \gls{dem} (HMADEM, SRTM)
        \end{itemize}
    } & 
    \Block[c]{1-1}{\gls{hma}: Qilian} & 
    \Block[c]{1-1}{-} \\

    \midrule

    \Block{1-1}{\citet{thomas_integrated_2023}}  & 
    \Block{1-1}{custom (5-layers \gls{cnn})} &
    \Block{1-1}{
        \begin{itemize}[leftmargin=0.25cm, itemsep=-0.4ex]
             \item optical (Sentinel-2)
            \item \gls{insar} coherence (Sentinel-1)
            \item \gls{dem} (AW3D30)
            \item thermal (Landsat-8)
        \end{itemize}
    } & 
    \Block[c]{1-1}{\gls{hma}: Khumbu, Manaslu, Hunza} & 
    \Block[c]{1-1}{-} \\

    \midrule
    
    \Block{1-1}{\citet{maslov_glavitu_2023} (GlaViTU)} & 
    \Block{2-1}{SETR \& U-Net \citep{zheng_rethinking_2021, navab_u-net_2015}} &
    \Block[l]{2-1}{ 
        \begin{itemize}[leftmargin=0.25cm, itemsep=-0.4ex]
            \item optical (Landsat 7/8, Sentinel-2)
            \item \gls{sar}\tablefootnote{In both studies $\sigma_0$-calibrated amplitude images are used. In the second study, \acrshort{insar} coherence images are also included when available.} (Envisat, Sentinel-1)
            \item \gls{dem} (AW3D30, Cop30DEM, SRTM)
            \item thermal\tablefootnote{used only in the second study} (Landsat-8)
        \end{itemize}
        } & 
    \Block[c]{1-1}{European Alps, \acrshort{hma}, Indonesia, New Zealand, S Andes and Scandinavia} & 
    \Block{1-1}{\href{https://github.com/konstantin-a-maslov/GlaViTU-IGARSS2023}{C}/\href{https://drive.google.com/drive/folders/1ivi18vaGXdsaxbJLKAt9R8rzOUvgg1Em}{D}/-} \\

    \cmidrule{1-1}\cmidrule{4-5}

    \Block{1-1}{\citet{maslov_towards_2024} (GlaViTU\tablefootnote{In this work, the original GlaViTU model from \citet{maslov_glavitu_2023} is slightly modified, see \Cref{sec:glacier-extent-mapping-standard}})} & 
    &
    & 
    \Block[c]{1-1}{global} & 
    \Block{1-1}{\href{https://github.com/konstantin-a-maslov/towards_global_glacier_mapping}{C}/\href{https://drive.google.com/drive/folders/1wu_DTCknr5ozu5e8FYqRlMLcmNGbOsxA}{D}/-} \\

    \bottomrule
    
    \end{NiceTabular}
    \caption{Summary of \gls{dl}-based studies for standard glacier-extent mapping. For details, see \Cref{sec:glacier-extent-mapping-standard} or the corresponding publications. The full links are also provided in \Cref{sec:resources}.}
    
    \label{tab:glacier-extent-mapping-standard}
\end{table}

\subsubsection{Area Change Estimation}
\label{sec:glacier-extent-mapping-area-change}

\begin{figure}
\centering
    \centering
    \includegraphics[trim={0.0cm 0.0cm 0.0cm 0.0cm}, clip, height=0.3\textheight]{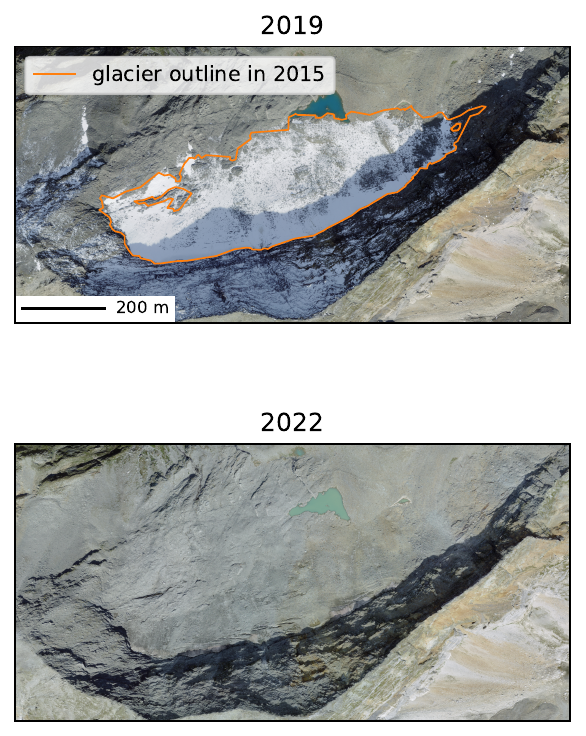}
    \captionof{figure}{\textbf{Debris-covered glacier with fresh snow}. The effect of fresh snow when mapping a fully debris-covered glacier is here shown for Glatscher da Sut Fuina, in Switzerland (46.535$^{\circ}$ N, 9.473$^{\circ}$ E) (same glacier as in \Cref{fig:dhdt}). The two aerial images from \citet{swisstopo_swissimage_2024} capture an important issue that can affect automatic glacier extent mapping \& area change analysis, i.e. predicting the presence of a glacier using, as a proxy, the superimposed fresh snow, hardly present outside the glacier surface (upper panel, where the snow only remains over the debris-covered glacier, which is colder than the non-glaciated surroundings). Since automatic methods, including \gls{dl}-based ones, are usually challenged in the case of debris cover, in this example, one would probably underestimate the glacier surface in 2022 (without snow) and thus overestimate the shrinkage rate. This figure illustrates that choosing imagery with similar climatic conditions, ideally without any snow, should be a priority, which is especially important for glacier area change analyses. See also \Cref{fig:gl-area-change-issue-2}.}
    \label{fig:gl-area-change-issue-1}
\end{figure}

\begin{figure}
\centering
    \centering
    \includegraphics[trim={0.0cm 0.0cm 0.0cm 0.0cm}, clip, height=0.4\textheight]{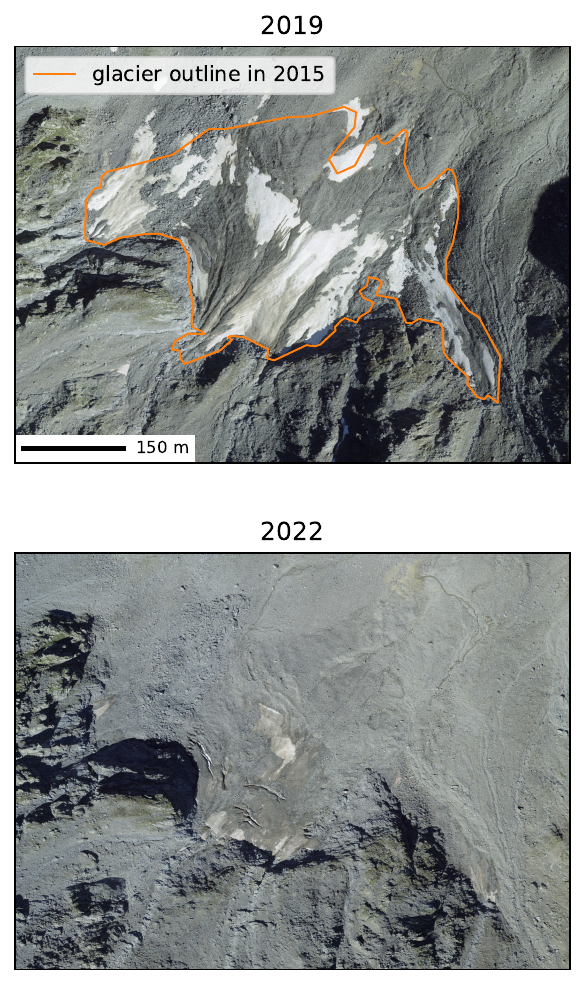}
    \captionof{figure}{\textbf{A glacier with increasing debris coverage}. For many debris-covered glaciers, there is a tendency for debris cover to increase over time, as illustrated here for the Tambogletscher, a glacier in Switzerland (46.504$^{\circ}$ N, 9.291$^{\circ}$ E). The two aerial images from \citet{swisstopo_swissimage_2024} show that initially, in 2019, the glacier was only partially covered by debris (upper panel), while three years later (lower panel), it has become almost completely debris-covered. Such transitions will affect automatic glacier extent mapping \& area change analysis since automatic methods, including \gls{dl}-based ones, usually face important challenges related to detecting the presence of debris. In the example presented here, many detection methods will likely underestimate the glacier area in 2022 and thus overestimate the glacier shrinkage rate. This figure suggests that capturing the uncertainties in the methods becomes critical to avoid significant biases in the estimates. See also \Cref{fig:gl-area-change-issue-1}.}
    \label{fig:gl-area-change-issue-2}
\end{figure}

Since the area of a glacier is a crucial indicator of its health, tracking its evolution over time is an important application of glacier mapping methods. Traditionally, this is done by manually re-creating glacier inventories after a certain period, usually decades, such that long-term impacts of climate change can be observed (see  \Cref{sec:intro-gl-mapping} where we provide examples). In principle, once a \gls{dl} model is trained (as presented in the previous section), it can be applied to delineate glaciers at different points, with the predicted results being used to analyze the temporal changes. However, a few challenges arise in practice when detecting glacier changes from outlines derived from a \gls{dl} framework, requiring additional assumptions. To list a few: 
\begin{itemize}
    \item To extract an area change rate for each individual glacier, the same ice divides have to be used, as usually the methods only classify a pixel as being part of the glacier or not, without any knowledge about what a glacier is as an independent object. This can be problematic when tributaries are becoming disconnected from the main glacier.  
    \item In general, to increase the signal-to-noise ratio, there should be a relatively long period between the considered outlines. However, most models are trained on imagery from a single source, e.g., Landsat-8, usually constrained by the static inventory that provides the labels, e.g., the \gls{rgi}. If the resulting model is then applied on data from different sensors to increase the temporal coverage, we potentially need to deal with generalization issues. 
    \item Even if the data comes from the same sensor, temporal generalization issues can still occur. One such situation is when the imagery has different characteristics compared to the one used for training, e.g. different cloud coverage or seasonal snow conditions. See \Cref{fig:gl-area-change-issue-1} as an example.
    \item Increasing debris-coverage in a warming climate \citep{tielidze_supra-glacial_2020, compagno_modelling_2022} can also affect the temporal generalization. Since automatic methods are still significantly more affected by errors on debris-covered glacier parts than clean ice, the change in the debris coverage percentage can introduce biases in the results. An example is provided in \Cref{fig:gl-area-change-issue-2}.
    \item Once a model is trained, we ideally want to employ it on the entire \gls{roi} to cover all the glaciers. However, this requires applying the model also on the data on which it was trained. Given the risk of memorization, especially when training with noisy labels \citep{arpit_closer_2017}, we can expect that the model errors over time are not independent, thus breaking the temporal generalization assumption. 
\end{itemize}

To summarize, additional challenges arise when relying on \gls{dl} for glacier area change quantification. These challenges perhaps explain why there are only a relatively small number of \gls{dl} works that explicitly focus on glacier changes. A few works that attempted to identify glacier changes are presented in the following three paragraphs and summarized in \Cref{tab:glacier-extent-mapping-area-change}.

\paragraph*{GlacierCoverNet \citep{roberts-pierel_changes_2022}} The availability of long time-series data from the Landsat program offers some opportunities for glacier studies. \citet{roberts-pierel_changes_2022} exploit the data starting from 1985 up to 2020 covering Alaska to study how glacier surface area evolved over time, an important indicator of the glaciers' health. After building a temporal mosaic to account for clouds and seasonal snow, they obtained 18 biannual image composites, each with full spatial coverage of the selected region. Based on this, they extract five features: \gls{ndsi}, \gls{ndvi}, \gls{nbr} and Tasseled Cap Brightness \& Wetness \citep{kauth_tasselled_1976}. Additionally, a \gls{dem} is used as input, together with three features derived from it: curvature and aspect intensity (North and South), i.e. a method of scaling the cosine of aspect by the sine of slope \citep{kirchner_lidar_2014}. The proposed model is based on the FSPNet architecture \citep{zhao_pyramid_2017} with a ResNeSt-101  backbone \citep{zhang_resnest_2022}. The glacier outlines from \gls{rgi} 6.0 were used for training. Another significant difference compared to the previous studies is that the model is trained to predict separately whether a pixel is no glacier, supra-glacial debris or debris-free glacier. This was achieved by developing a method for identifying the debris pixels and then assuming that those within the \gls{rgi} outlines are debris-covered glacier pixels. Once the model was trained using the image composites as close as possible to the \gls{rgi} dates, it was applied on the entire time-series, thus producing 18 glacier inventories for Alaska. These show a significant area loss: the total glaciated area decreased by 8425 km$^2$ (i.e. -13\%), with sizeable sub-regional variability. The relatively fine temporal resolution also allows the inspection of the glacier area changes through time, showing a much stronger shrinkage over the last 15 years (2005-2020) compared to the earlier period. Lastly, the distinction between debris and clean ice allows studying the debris evolution over time, revealing an increase of around 64\% over 1985-2020, also with significant sub-regional variability.

\paragraph*{\citet{rajat_glacier_2022}}
A similar investigation but on a much smaller scale shows that the Himachal glaciers retreated significantly, from an estimated total area of around 4,021 km$^2$ in 1994 to only 2199 km$^2$ in 2021, resulting in an annual retreat rate of $\sim$68 km$^2$ ($\sim$1.68\%). To obtain these results, a network was trained on manually annotated glaciers by visualizing Landsat-8-based \gls{ndsi} and a \gls{dem} from USGS, with additional derived features. The model, a U-Net \citep{navab_u-net_2015}, is trained using a subset of four bands and then applied on four different years, i.e.  1994, 2001, 2011 and 2021, using input data from Landsat 4/5/8.

\paragraph*{\citet{diaconu_detailed_2023}} This is another study that makes use of \gls{dl} for investigating glacier area changes, focused on a different region, the European Alps (\gls{rgi}-11). The input data consists of a subset of five bands (R, G, B, \gls{nir}, \gls{swir}-B12) from  Sentinel-2 and a \gls{dem}. A major advantage of this study is the relatively good quality of the labels: they are based on a new inventory \citep{paul_glacier_2020} and estimated to be of better quality than the previous \gls{rgi} one. It is based on Sentinel-2 data from (mainly) 2015, ensuring a perfect match between the satellite data used for training and the glacier outlines. The model used is a U-Net \citep{navab_u-net_2015} with a ResNet34  backbone \citep{he_deep_2016}. To avoid using inferences made on the training data, which can lead to biases, five models are trained using a regional cross-validation scheme. Once the models are trained, they are applied on the most recent data, i.e. 2023, which was a strong melt year according to \gls{glamos} \citep{glamos_figures}. This offers, however, ideal conditions for glacier mapping as it reduces the chances of seasonal snow, which can cause many false positives. Based on the models' predictions, the areas are estimated both at the inventory time and in 2023, which are then compared to estimate change rates for each individual glacier. To increase the signal-to-noise ratio, the authors make two important assumptions: i) the glaciers do not grow in the given period, supported by geodetic mass-balance studies \citep{hugonnet_accelerated_2021} and ii) the models make systematic errors. The second assumption, if true, implies that the segments missed by the models, e.g. debris-covered or under shadow, do not significantly affect the estimated change rates. The regional cross-validation scheme helps support this assumption since it decreases the chances that models would perform differently at the inventory time compared to 2023. To further reduce the impact of the models' errors on the area change rates, an outlier filtering scheme was used to drop the glaciers for which the model performs poorly. After this step, individual estimates for around 1,300 glaciers are provided, representing 87\% of the glacierized area in the region. Regionally, the estimate is around -1.8\% loss per year, which illustrates the high sensitivity of the glaciers in this region to climate change. A glacier-level analysis further shows significant inter-glacier variability. \\

\begin{table}[!ht]
    \centering
        \begin{NiceTabular}{|m{0.12\textwidth}|m{0.2\textwidth}|m{0.25\textwidth}|m{0.15\textwidth}|m{0.08\textwidth}|}
    \toprule
    \RowStyle[bold]{}
    
    \Block{1-1}{Publication (model name)} &
    \Block{1-1}{model architecture\tablefootnote{Note that sometimes changes to the original architecture are made, see \Cref{sec:glacier-extent-mapping-standard}.}} &
    \Block{1-1}{data modality\tablefootnote{Often features are further derived e.g. \gls{ndvi} from optical or slope from \gls{dem}.} (source)} &
    \Block{1-1}{\acrshort{roi}\tablefootnote{We only indicate the main \glspl{roi} from where the training data was extracted. Still, the coverage can vary significantly, e.g. from a few image scenes to almost complete coverage.} } &
    \Block{1-1}{code/ data\tablefootnote{This refers to the processed training data, not the raw one. The latter is usually openly available from the original source.}/ output\tablefootnote{In case the study has an output product (e.g. an inventory).}}  \\
    
    \midrule

    \Block{1-1}{\citet{roberts-pierel_changes_2022} (GlacierCoverNet)}  & 
    \Block{1-1}{FSPNet \\ \citep{zhao_pyramid_2017}} &
    \Block{1-1}{
        \begin{itemize}[leftmargin=0.25cm, itemsep=-0.4ex]
            \item optical (Landsat 4/5/7/8)
            \item \gls{dem} (USGS-3DEP)
        \end{itemize}
    } & 
    \Block[c]{1-1}{Alaska} &
    \Block[c]{1-1}{-/-/\href{https://doi.org/10.7265/8esq-w553}{O}} \\

    \midrule
    
    \Block{1-1}{\citet{rajat_glacier_2022}}  & 
    \Block{1-1}{U-Net \\ \citep{navab_u-net_2015}} &
    \Block{1-1}{
        \begin{itemize}[leftmargin=0.25cm, itemsep=-0.4ex]
            \item optical (Landsat 4/5/8)
        \end{itemize}
    } & 
    \Block[c]{1-1}{\gls{hma}: Himachal} & 
    \Block[c]{1-1}{-}\\

    \midrule

    \Block{1-1}{\citet{diaconu_detailed_2023}}  & 
    \Block{1-1}{U-Net \\ \citep{navab_u-net_2015}} &
    \Block{1-1}{
        \begin{itemize}[leftmargin=0.25cm, itemsep=-0.4ex]
            \item optical (Sentinel-2)
            \item \gls{dem} (NASADEM)
        \end{itemize}
    } &
    \Block[c]{1-1}{European Alps (\gls{rgi}-11)} & 
    \Block[c]{1-1}{\href{https://github.com/dcodrut/glacier_mapping_alps_tccml}{C}/\href{https://huggingface.co/datasets/dcodrut/glacier_mapping_alps}{D}/-} \\

    \bottomrule
    
    \end{NiceTabular}
    \caption{Summary of \gls{dl}-based studies focused on glacier area change analysis. For details, see \Cref{sec:glacier-extent-mapping-area-change} or the corresponding publications. The full links are also provided in \Cref{sec:resources}.}
    \label{tab:glacier-extent-mapping-area-change}
\end{table}

\subsubsection{Rock Glaciers Mapping}
\label{sec:glacier-extent-mapping-rock-glaciers}

\begin{figure}
\centering
    \centering
    \includegraphics[trim={0.0cm 0.0cm 0.0cm 0.0cm}, clip, width=0.7\textwidth]{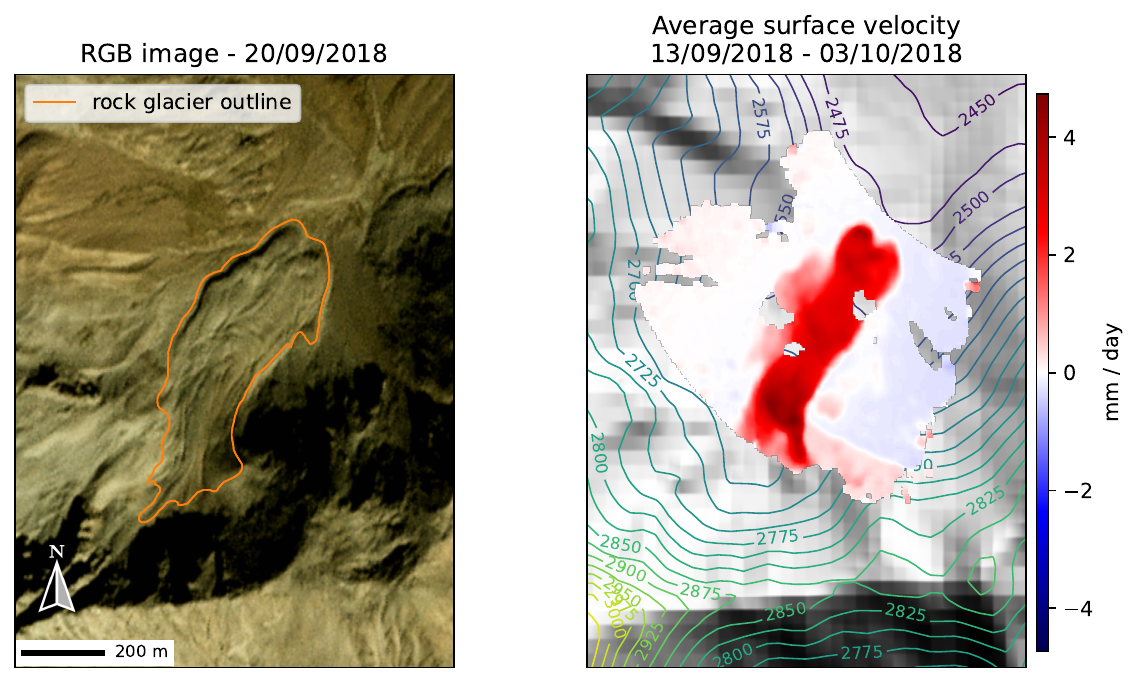}
    \captionof{figure}{\textbf{An active rock glacier}. Some rock glaciers that have substantial amounts of ice are subject to a distinct downslope (gravitational) flow, as illustrated here for the Lazaun glacier in northern Italy (46.746$^{\circ}$ N, 10.755$^{\circ}$ E). The satellite image (left panel) from PlanetScope (3 m \gls{gsd}) shows the extent of the glacier, which is difficult to distinguish from the surrounding landforms. On the right panel, we show the average daily surface velocity from \citet{bertone_unprecedented_2023} measured using ground-based \gls{sar} (with Copernicus GLO-30 DEM in the background). This figure illustrates how similar a rock glacier is to the surrounding landscape and how utilizing surface velocity observations can contribute to a better distinction. }
    \label{fig:rock-glacier}
\end{figure}

A special class of glaciers is the so-called rock glaciers. These are essentially a mixture of frozen debris and ice\footnote{For a more technical definition, see \citet{rgik_guidelines_2023}.}. As opposed to typical glaciers, which, by definition, are flowing, rock glaciers can be both active and non-active. They are important for cryosphere studies as they can indicate the permafrost distribution in the region. Rock glaciers are also affected by climate change but are more resilient due to the insulated effect of the rocky material and the active layer \citep{robson_automated_2020}. Detecting this type of glaciers from optical data is even more difficult as, by definition, they are covered by debris. Consequently, the studies on this area usually have a multi-modal approach, e.g., by including \gls{sar} coherence that can capture small deformations that could occur over time (at least for the active ones). An example of a rock glacier is provided in \Cref{fig:rock-glacier}. In the following paragraphs, we describe three major studies focused on rock glaciers and then summarized in \Cref{tab:glacier-extent-mapping-rock-glaciers}.

\paragraph*{\citet{robson_automated_2020}} 
Rock glaciers pose even more challenges to automated methods compared to glaciers that are only partially covered by debris due to the spectral similarity between them and the surrounding material. \citet{robson_automated_2020} propose to use a \gls{cnn} model combined with \gls{obia} to further improve the predictions using additional morphological and spatial characteristics. Given the difficulty of the task, various data sources are used: optical data from Sentinel-2, \gls{insar} coherence data from Sentinel-1 and a Pléiades \gls{dem}  (processed by the authors). The study focuses on the La Laguna and Poiqu catchments, in the Andes, and the central Himalayas, respectively. The ground truth labels are obtained from a dataset by Schaffer and Macdonell for the La Laguna catchment and the one from \citet{bolch_occurrence_2020} for Poiqu.
The authors used a relatively small \gls{dl} model, a five-layers \gls{cnn} trained with eleven input features: five optical bands (R, G, B, \gls{nir}, \gls{swir}) + three derived indices (\gls{ndvi}, \gls{mndwi}\citep{xu_modification_2006}, \gls{savi} \citep{alba_automatic_2012}), \gls{insar} coherence, \gls{dem} + derived curvature. With the full pipeline, i.e. \gls{cnn} + \gls{obia}, they automatically mapped 108 of the 120 glaciers considered in the validation set. The authors also investigated, for a small sub-region of the Poiqu catchment, whether using higher resolution optical data and the corresponding \gls{dem} from Pléiades (2m) boosts the performance and found a 9\% increase in precision but only 1\%  increase in recall. 

\paragraph*{\citet{hu_mapping_2023}} 
If there are already many inventories for typical glaciers, including \gls{rgi} \citep{pfeffer2014randolph}, which has global coverage, rock glaciers are yet to be identified in some regions. The Western Kunlun Mountains represent one such case. \citet{hu_mapping_2023} built an inventory for the active rock glaciers in the region using \gls{insar} from ALOS and \gls{vhr} images from Google Earth. Using this data, a DeepLabv3+ model \citep{chen_encoder-decoder_2018} with an Xception backbone \citep{chollet_xception_2017} was trained on Sentinel-2 (only RGB) and applied to the entire region to further identify glaciers that were previously missed, increasing the initial number of 290 glaciers to 413. The outlines produced by the model had however to be manually inspected and corrected, therefore further research is necessary towards a fully-automatic pipeline. 

\paragraph*{\citet{sun_tprogi_2024}}
For the first time, a regional-scale inventory for rock glaciers was built by \citet{sun_tprogi_2024}. The final benchmark dataset contains 44,273 glaciers covering $\sim$6000 km$^2$ ($\mu$ = 0.14 km$^2$). This was achieved by initially compiling multiple existing inventories, not only from the Tibetan Plateau but also from other regions, into a large dataset containing 4085 rock glaciers. This was then used to train a \gls{dl} model using Planet Basemaps. The model architecture is the same as in \citet{hu_mapping_2023}, i.e. a DeepLabv3+ model \citep{chen_encoder-decoder_2018} with an Xception backbone \citep{chollet_xception_2017}. The initial predictions of the \gls{dl} model were investigated and manually corrected following a strict guideline \citep{rgik_guidelines_2023}, with the effort of seven mappers and two independent reviewers. At this stage, high-resolution Google Earth images and ESRI basemaps were also utilized. When comparing the initially \gls{dl}-produced polygons with the final ones revised by experts, a 63\% F1 score was obtained (precision = 55\%, recall = 73\%).

\begin{table}[!ht]
    \centering
        \begin{NiceTabular}{|m{0.12\textwidth}|m{0.2\textwidth}|m{0.25\textwidth}|m{0.15\textwidth}|m{0.08\textwidth}|}
    \toprule
    \RowStyle[bold]{}
    
    \Block{1-1}{Publication (model name)} &
    \Block{1-1}{model architecture\tablefootnote{Note that sometimes changes to the original architecture are made, see \Cref{sec:glacier-extent-mapping-standard}.}} &
    \Block{1-1}{data modality\tablefootnote{Often features are further derived e.g. \gls{ndvi} from optical or slope from \gls{dem}.} (source)} &
    \Block{1-1}{\acrshort{roi}\tablefootnote{We only indicate the main \glspl{roi} from where the training data was extracted. Still, the coverage can vary significantly, e.g. from a few image scenes to almost complete coverage.} } &
    \Block{1-1}{code/ data\tablefootnote{This refers to the processed training data, not the raw one. The latter is usually openly available from the original source.}/ output\tablefootnote{In case the study has an output product (e.g. an inventory).}}  \\
    
    \midrule
    \Block{1-1}{\citet{robson_automated_2020}}  & 
    \Block{1-1}{custom (5-layers \gls{cnn})} &
    \Block{1-1}{
        \begin{itemize}[leftmargin=0.25cm, itemsep=-0.4ex]
            \item optical (Sentinel-2, Pléiades)
            \item \gls{insar} coherence (Sentinel-1)
            \item \gls{dem} (generated from Pléiades)
        \end{itemize}
    } &
    \Block[c]{1-1}{La Laguna catchment (Andes) \\ Poiqu catchment (Himalaya)} & 
    \Block[c]{1-1}{-}
    \\

    \midrule

    \Block{1-1}{\citet{hu_mapping_2023}}  & 
    \Block{1-1}{DeepLabv3+ \citep{chen_encoder-decoder_2018}} &
    \Block{1-1}{
        \begin{itemize}[leftmargin=0.25cm, itemsep=-0.4ex]
            \item optical (Sentinel-2)
        \end{itemize}
    } &
    \Block[c]{1-1}{\gls{hma}: Western Kunlun} & 
    \Block[c]{1-1}{\href{https://zenodo.org/records/8223732}{C}/-/\href{https://doi.pangaea.de/10.1594/PANGAEA.938686}{O}}
    \\

    \midrule

    \Block{1-1}{\citet{sun_tprogi_2024}}  & 
    \Block{1-1}{DeepLabv3+ \citep{chen_encoder-decoder_2018}} &
    \Block{1-1}{
        \begin{itemize}[leftmargin=0.25cm, itemsep=-0.4ex]
            \item optical \\(Planet Basemaps)
        \end{itemize}
    } &
    \Block[c]{1-1}{\gls{hma}: Tibetan Plateau} & 
    \Block[c]{1-1}{-/-/\href{https://zenodo.org/records/10732042}{O}}
    \\

    \bottomrule
    
    \end{NiceTabular}
    \caption{Summary of \gls{dl}-based studies focused on rock glaciers mapping. For details, see \Cref{sec:glacier-extent-mapping-rock-glaciers} or the corresponding publications. The full links are also provided in \Cref{sec:resources}.}
    \label{tab:glacier-extent-mapping-rock-glaciers}
\end{table}


\FloatBarrier
\subsection{Calving Front Detection}
\label{sec:cf-detection}
For marine-terminating glaciers, changes in the calving front are an essential indicator of the underlying glacial dynamics, with shifts in the calving front hinting at melt processes or surge events.
As the mass loss of the major ice sheets in Antarctica and Greenland is projected to be a major contributor to global sea level rise~\citep{calvin2023_ipcc}, understanding and monitoring these developments is paramount.

Various \gls{dl} approaches have been suggested in recent years to automatically monitor calving fronts for the Earth's ice sheets.
Generally, these methods use either optical or \gls{sar} imagery as their primary source of imagery, with some methods taking auxiliary data as additional inputs, such as elevation data.
The following is an overview of relevant existing works on \gls{dl} for calving front detection:

\paragraph*{\citet{baumhoer2019_automated}}
Seeing the potential of \gls{dl} methods for automated detection of ice sheet calving fronts in Antarctica, \citet{baumhoer2019_automated} adapted the U-Net model \citep{navab_u-net_2015} for this task.
As a main input feature, dual-polarized Sentinel-1 \gls{sar} data was chosen for its year-round availability and robustness against weather conditions.
For training, the calving front was manually annotated in \gls{sar} imagery for the Sulzberger, Victoria Land, Wilkes Land and Shackleton regions.
Four additional regions were selected for testing purposes: Ekstromisen, Wordie, Marie Byrd Land and Oats Land.
As an additional input feature for the model, the HH/HV ratio was derived. 
Finally, TanDEM-X elevation data was merged into the data stack as a fourth input to guide the model in regions with confounding backscatter behaviour.

\paragraph*{\citet{mohajerani2019_detection}}
In a first case study for Greenland glaciers, \citet{mohajerani2019_detection} trained a U-Net model \citep{navab_u-net_2015} to detect calving fronts for three glaciers in Landsat imagery.
For Landsat 5, the green band was extracted, while the panchromatic band was used for Landsat 7 and 8.
Calving fronts were annotated manually for each image.
The studied regions cover the Helheim, Sverdrup, Kangerlussuaq, and Jakobshavn glaciers.
The imagery is segmented into two classes, namely a ``background'' class and a ``calving front'' class, following an edge-detection approach rather than a zonal segmentation approach.
The final calving front prediction is then derived via a path-finding algorithm and re-projection to the original geographical coordinates.

\paragraph*{\citet{zhang2019_automatically}}
Seeing the value of \gls{sar} data for calving front detection, \citet{zhang2019_automatically} manually delineated coastline positions for the Jakobshavn Isbrae glacier in Greenland using imagery from the TerraSAR-X mission.
As a segmentation model, they train a U-Net model~\citep{navab_u-net_2015} to segment the imagery into a ``glacier'' and an ``ocean'' class.
The calving front is then extracted by tracing the boundary between these two classes in a post-processing step.

\paragraph*{CALFIN \citep{cheng2021_calving}}
\citet{cheng2021_calving} built a large-scale dataset of calving front observations in Greenland by manually annotating Landsat data from 1972 to 2019, covering 66 glaciers.
Similar to \citet{mohajerani2019_detection}, single-channel data is used (in this case, from the near-infrared band).
However, the imagery is pre-processed into three channels by applying contrast normalization algorithms and stacking the results.
Using this dataset, a DeepLabv3+ model \citep{chen_encoder-decoder_2018} with an Xception backbone \citep{chollet_xception_2017} is then trained to both segment the imagery into ocean and land classes, as well as directly mark the calving front.
In a post-processing step, the calving front is then extracted by constructing a minimum-spanning tree and finding the longest path within this tree.

\paragraph*{\citet{zhang2021_automated}}
Following up on their previous study~\citep{zhang2019_automatically} extend their dataset to span three glaciers, namely Jakobshavn, Kangerlussuq, and Helheim.
Further, multiple data modalities are included, namely optical imagery from Landsat-8 and Sentinel-2, as well as \gls{sar} imagery from Envisat, ALOS-1, TerraSAR-X, Sentinel-1 and ALOS-2.
Multiple models are trained to perform a binary segmentation into land/glacier and ocean, with the best performance observed for a DeepLabv3+ model~\citep{chen_encoder-decoder_2018} with a DRN backbone~\citep{yu2017_dilated}.

\paragraph*{HED-UNet~\citep{heidler2022_hedunet}}
Seeing the advantages of both segmentation and edge detection approaches for the task of calving front detection, \citet{heidler2022_hedunet} combined these two tasks in a single model.
This dual training enhances model predictions near the calving front, which tend to be imprecise and blurry for segmentation-based models.
The developed model takes inspiration from both the U-Net~\citep{navab_u-net_2015} and the HED edge detector~\citep{xie2015_holisticallynested}.
The dataset from \citet{baumhoer2019_automated} is used as training and evaluation data.
However, the DEM input channel is identified as a potential confounder in this study, which can cause the model to overfit to this static data product and ignore the actual \gls{sar} imagery.
The authors, therefore, advocate against using elevation data as a direct input feature for the \gls{dl} model and instead suggest only incorporating this data in a separate post-processing step.
IceLines~\citep{baumhoer2023_icelines} is a continuously updated data product for the entire Arctic derived using this model.

\paragraph*{\citet{loebel2022_extracting}}
Seeing that previous studies for calving front extraction in Greenland mostly relied on single-channel imagery, \citet{loebel2022_extracting} studied the benefits of incorporating various additional data modalities into the training process.
Using Landsat-8 imagery as the main imagery source, they manually delineate calving fronts for 23 glaciers in Greenland glaciers and two glaciers on the Antarctic Peninsula.
The used \gls{dl} model is a deepened version of the U-Net~\citep{navab_u-net_2015} model, including two additional down- and up-sampling stages to allow for a larger spatial context.
Starting with panchromatic imagery as a baseline input modality, the authors test various configurations by adding multi-spectral channels, statistical texture information, and topography data from the BedMachine Greenland v3 dataset~\citep{morlighem2017_bedmachine}.
The largest gains in model accuracy are observed when multi-spectral information is included.
Meanwhile, the inclusion of textural and topographic information seems to introduce a trade-off, where the model performance becomes more robust on challenging scenes, but, in turn, accuracy decreases for most other images.

\paragraph*{\citet{gourmelon2022_calving}}
In an effort to make the task of calving front detection more approachable for researchers from the field of computer vision, \citet{gourmelon2022_calving} build CaFFe, a ``machine-learning ready'' dataset of calving front positions in Greenland, Antarctica, and Alaska.
The imagery used comes from various \gls{sar} sensors, namely ERS-1/2, RADARSAT 1, Envisat, ALOS, TerraSAR-X, TanDEM-X, and Sentinel-1.
As ground truth annotations, the authors provide both calving front masks for edge-oriented approaches and zone labels that segment the imagery into the classes ``ocean'', ``rock'', ``glacier'', and ``NA''.
As a baseline model for future comparisons, the authors train an adapted U-Net model~\citep{navab_u-net_2015}, which is augmented by an atrous spatial pyramid pooling layer~\citep{chen_encoder-decoder_2018} in the bottleneck of the network.
Notably, this study quantifies the effect of the seasons and image resolution on the prediction accuracy, suggesting that summer images are to be preferred over winter images and that higher resolution can help with prediction accuracy in some cases.

\paragraph*{\citet{periyasamy2022_how}}
Seeing the wide-spread use of the U-Net model~\citep{navab_u-net_2015} for calving front detection, \citet{periyasamy2022_how} conducted a systematic study to better understand the influence of specific hyperparameters for this task and give recommendations on how to tune models for calving front detection.
Using an earlier version of the CaFFe dataset~\citep{gourmelon2022_calving}, they optimize various components of the training process, such as data preprocessing, data augmentation, loss function, bottleneck, normalization layers and dropout layers.
They observe optimal performance when using adaptive histogram equalization.

\paragraph*{COBRA~\citep{heidler2023_deep}}
Nearly all existing studies employing \gls{dl} for calving front detection are trained to provide dense predictions, either in the form of semantic segmentation or edge detection.
\citet{heidler2023_deep} take another approach:
by adopting the idea of deep active contours~\citep{peng2020_deep}, they introduce a model that directly predicts a contour line, parameterized by a sequence of vertices in the image space.
In this way, the model is encouraged to focus on the actual calving front during training, and the predictions can be used without any post-processing steps.
This model is trained on the CALFIN dataset~\citep{cheng2021_calving}, and has been applied for a large-scale study of calving front dynamics in Svalbard~\citep{li2024_highresolution}.

\paragraph*{\citet{zhang2023_autoterm}}
To leverage the increasing amount of openly available remote sensing data, \citet{zhang2023_autoterm} developed an automated pipeline for calving front extraction using both \gls{sar} (from Sentinel-1) and optical data ( Landsat 5,7,8 and Sentinel-2). For training labels, the authors make use of TermPicks \citep{goliber_termpicks_2022}, a large-scale dataset with manually digitized calving fronts. To quantify the uncertainties in the model (a DeepLabv3+ \citep{chen_encoder-decoder_2018}), Monte Carlo dropout~\citep{gal_dropout_2016} is employed, combined with a temporal ensemble (i.e. combining multiple predictions for the same date). The high-temporal resolution of their results allows them to capture the seasonal variability, with the final product, AutoTerm, covering 295 outlet glaciers in Greenland and  278,239 calving fronts.

\paragraph*{AMD-HookNet~\citep{wu2023_amdhooknet}}
A common observation in calving front detection research is the need for large spatial context windows.
Naive solutions to addressing this requirement, such as training on larger image patches and increasing the size of convolutional filters, are computationally inefficient.
Setting out to address this issue in a more elegant manner, \citet{wu2023_amdhooknet} introduce AMD-HookNet, a deep neural network designed to operate on two versions of a satellite scene at different spatial resolutions.
By interlocking two U-Net \citep{navab_u-net_2015} branches with attention layers, more general information with a wide spatial context can be applied to improve the predictions of the high-resolution branch.
This model is trained on the CaFFe dataset introduced by \citet{gourmelon2022_calving}.
Evaluations show that this dual-resolution approach can indeed improve prediction accuracy considerably.

\paragraph*{HookFormer~\citep{wu2024_contextual}}
Recently, vision transformers \citep{dosovitskiy_image_2020} have become the tool of choice for many computer vision tasks.
Following the ideas introduced in the previous study, \citet{wu2024_contextual} combined a Swin Transformer model~\citep{liu_swin_2021} with the HookNet approach, where a high-resolution and a low-resolution branch are interleaved.
The resulting model is again trained on the CaFFe dataset~\citet{gourmelon2022_calving}.
Compared to the previous evaluations, this transformer-based model outperforms the previous \gls{cnn}-based models.

\FloatBarrier
\begin{table}[!ht]
    \centering
        \begin{NiceTabular}{|m{0.15\textwidth}|m{0.23\textwidth}|m{0.25\textwidth}|m{0.12\textwidth}|m{0.08\textwidth}|}
    \toprule
    \RowStyle[bold]{}
    
    \Block{1-1}{Publication (model name)} &
    \Block{1-1}{model architecture\tablefootnote{Note that sometimes changes to the original architecture are made, see \Cref{sec:cf-detection}.}} &
    \Block{1-1}{[dataset\tablefootnote{Indicates whether a `benchmark` dataset is used.}]\\data modality\tablefootnote{Often features are further derived e.g. \gls{ndvi} from optical or slope from \gls{dem}.} (source)} &
    \Block{1-1}{\acrshort{roi}\tablefootnote{We only indicate the main \glspl{roi} from where the training data was extracted. Still, the coverage can vary significantly, e.g. from a few image scenes to almost complete coverage. We provide the number of glaciers, where possible, but this should only be taken as a rough estimate as it depends on how glacier systems are separated into individual glaciers.} } &
    \Block{1-1}{code/ data\tablefootnote{This refers to the processed training data, not to the raw one. The latter is usually openly available from the original source. $\uparrow$ means see the data link from the previous publication.}/ output\tablefootnote{In case the study has an output product (e.g. an inventory).}}  \\
    
    \midrule

    \Block{1-1}{\citet{baumhoer2019_automated}}  & 
    \Block{1-1}{U-Net \citep{navab_u-net_2015}} &
    \Block{2-1}{
        \begin{itemize}[leftmargin=0.25cm, itemsep=-0.4ex]
            \item \gls{sar} (Sentinel-1)
            \item \gls{dem} (TanDEM-X)
        \end{itemize}
    } & 
    \Block[c]{2-1}{Antarctica \\ (8 sites)} &
    \Block[c]{1-1}{-} \\

    \cmidrule{1-2}\cmidrule{5-5}

    \Block{1-1}{\citet{heidler2022_hedunet} (HED-UNet)} &
    \Block{1-1}{custom; based on HED \citep{xie2015_holisticallynested} \& U-Net \citep{navab_u-net_2015} } & 
    &
    & 
    \Block{1-1}{\href{https://github.com/khdlr/HED-UNet}{C}/-/\href{https://geoservice.dlr.de/web/maps/eoc:icelines}{O}} \\
    
    \midrule

    \Block{1-1}{\citet{mohajerani2019_detection}} &
    \Block{1-1}{U-Net \citep{navab_u-net_2015}} &
    \Block{1-1}{
        \begin{itemize}[leftmargin=0.25cm, itemsep=-0.4ex]
            \item optical (Landsat 5,7,8)
        \end{itemize}
    } & 
    \Block{1-1}{Greenland\\(4 glaciers)} &
    \Block{1-1}{\href{https://github.com/yaramohajerani/FrontLearning/}{C}/\href{https://github.com/yaramohajerani/FrontLearning/tree/master/data/greenland_training.dir/data}{D}/-} \\

    \midrule

    \Block{1-1}{\citet{zhang2019_automatically}} &
    \Block{1-1}{U-Net \citep{navab_u-net_2015}} & 
   \Block{1-1}{
        \begin{itemize}[leftmargin=0.25cm, itemsep=-0.4ex]
            \item \gls{sar} (TerraSAR-X)
        \end{itemize}
    } & 
    \Block{1-1}{Jakobshavn Isbrae\\(Greenland)} &
    \Block{1-1}{-/-/\href{https://doi.pangaea.de/10.1594/PANGAEA.897066}{O}} \\
    
    \midrule

    \Block{1-1}{\citet{zhang2021_automated}} &
    \Block{1-1}{DeepLabv3+ \\ \citep{chen_encoder-decoder_2018}} & 
   \Block{1-1}{
        \begin{itemize}[leftmargin=0.25cm, itemsep=-0.4ex]
            \item optical (Landsat-8,\\  Sentinel-2)
            \item \gls{sar} (TerraSAR-X, \\Envisat, ALOS-1/2, Sentinel-1)
        \end{itemize}
    } & 
    \Block{1-1}{Greenland\\(3 glaciers)} &
    \Block{1-1}{\href{https://github.com/enzezhang/FrontDL3}{C}/\href{https://doi.pangaea.de/10.1594/PANGAEA.923270}{D}/\href{https://doi.pangaea.de/10.1594/PANGAEA.923272}{O}} \\

    \midrule

    \Block{1-1}{\citet{cheng2021_calving} (CALFIN)} &
    \Block{1-1}{DeepLabv3+ \\ \citep{chen_encoder-decoder_2018}} & 
    \Block{2-1}{
        CALFIN dataset \citep{cheng2021_calving}
        \begin{itemize}[leftmargin=0.25cm, itemsep=-0.4ex]
            \item optical \\ (Landsat 5,7,8)
        \end{itemize}
    } & 
    \Block{2-1}{Greenland\\(66 glaciers)} &
    \Block{1-1}{\href{https://github.com/daniel-cheng/CALFIN}{C}/\href{https://github.com/daniel-cheng/CALFIN/tree/master/training/data}{D}/\href{https://github.com/daniel-cheng/CALFIN/tree/master/outputs/upload_production/v1.0/level-1_shapefiles-domain-termini}{O}} \\

    \cmidrule{1-2}\cmidrule{5-5}
    
    \Block{1-1}{\citet{heidler2023_deep} (COBRA)} &
    \Block{1-1}{custom; based on Deep Snake \citep{peng2020_deep}} & 
     & 
     &
    \Block{1-1}{\href{ https://khdlr.github.io/COBRA/}{C}/$\uparrow$/\href{https://github.com/khdlr/COBRA/tree/master/inference_results}{O}} \\

    \midrule
   \Block{1-1}{\citet{loebel2022_extracting}} &
   \Block{1-1}{U-Net \citep{navab_u-net_2015}} & 
  \Block{1-1}{
        \begin{itemize}[leftmargin=0.25cm, itemsep=-0.4ex]
            \item optical (Landsat-8)
            \item bed topography (BedMachine Greenland v3 \citep{morlighem2017_bedmachine}) 
        \end{itemize}
    } & 
   \Block{1-1}{Greenland (23 glaciers) \& Antarctica (2 glaciers)} &
   \Block[c]{1-1}{\href{https://github.com/eloebel/glacier-front-extraction}{C}/\href{https://opara.zih.tu-dresden.de/xmlui/handle/123456789/5721}{D}/-} \\

    \midrule
   \Block{1-1}{\citet{periyasamy2022_how}} &
   \Block{1-1}{U-Net \citep{navab_u-net_2015}} & 
   \Block{1-2}{earlier version of CaFFe (see next line)} &
   & 
   \Block[c]{1-1}{-} \\
    
   \midrule
    
   \Block{1-1}{\citet{gourmelon2022_calving}} &
   \Block{1-1}{U-Net \citep{navab_u-net_2015}} & 
   \Block{3-1}{
        CaFFe dataset \citep{gourmelon2022_calving}:
        \begin{itemize}[leftmargin=0.25cm, itemsep=-0.4ex]
            \item \gls{sar} (ENVISAT, \\ ESR 1\&2, Sentinel-1, TerraSAR-X, TanDEM-X, ALOS, RADARSAT-1)
        \end{itemize}
    } & 
    \Block{3-1}{6 glaciers from Antarctica, Greenland, and Alaska} &
    \Block{1-1}{\href{https://github.com/Nora-Go/Calving_Fronts_and_Where_to_Find_Them}{C}/\href{https://doi.pangaea.de/10.1594/PANGAEA.940950}{D}/\href{https://zenodo.org/records/6469519}{O}} \\

    \cmidrule{1-2}\cmidrule{5-5} 

    \Block{1-1}{\citet{wu2023_amdhooknet} (AMD-HookNet)} &
    \Block{1-1}{custom; two branch U-Net \citep{navab_u-net_2015}} & 
    & 
    &
    \Block[c]{1-1}{\href{https://github.com/RiverNA/AMD-HookNet}{C}/$\uparrow$/-} \\
    
    \cmidrule{1-2}\cmidrule{5-5} 

    \Block{1-1}{\citet{wu2024_contextual} (HookFormer)} &
    \Block{1-1}{custom; two branch Swin Transformer \citep{liu_swin_2021}} & 
    & 
    &
    \Block[c]{1-1}{\href{https://github.com/RiverNA/HookFormer}{C}/$\uparrow$/-} \\

    \midrule

    \Block{1-1}{\citet{zhang2023_autoterm}} &
    \Block{1-1}{DeepLabv3+ \\ \citep{chen_encoder-decoder_2018}} & 
    \Block{1-1}{
        \begin{itemize}[leftmargin=0.25cm, itemsep=-0.4ex]
            \item \gls{sar} (Sentinel-1)
            \item optical (Sentinel-2, \\ Landsat-5,7,8)
        \end{itemize}
    } & 
    \Block{1-1}{Greenland\\(295 glaciers)} & 
    \Block{1-1}{\href{https://zenodo.org/records/8270875}{C}/-/\href{https://zenodo.org/records/7782039}{O}} \\

    \bottomrule
    
    \end{NiceTabular}
    \caption{Summary of \gls{dl}-based studies focused on calving-front detection. For details, see \Cref{sec:cf-detection} or the corresponding publications. The full links are also provided in \Cref{sec:resources}.}
    \label{tab:cf-detection}
\end{table}

\FloatBarrier
\section{Discussion}
\label{sec:discussion}

In this work, we provided an overview of glacier mapping with \gls{dl}, with a first part focusing on mapping (delineating) the full glacier extent (\Cref{sec:glacier-extent-mapping}) and a second part detailing the automatic extraction of calving fronts (\Cref{sec:cf-detection}). Although the two fields evolved relatively independently, the utilized methodologies are generally similar, with many works relying on fully convolutional segmentation models that are well established in the \gls{dl} community. 

Based on the data sources that these studies use, we can conclude that glacier mapping heavily relies on data fusion, as most of the methods make use of at least two different data modalities, e.g. for glacier extent mapping, usually, at least an optical and a \gls{dem} are combined. This multimodal characteristic indicates the advantage of using \gls{dl} for glacier mapping tasks, as it can automatically learn to extract useful information from each modality. We note, however, that most of the glacier mapping studies simply concatenate the input bands coming from each source, which may be sub-optimal (see, e.g. \citet{li_deep_2022} for a review on data fusion with \gls{dl}, focused on remote sensing applications). Some of the limitations we identified in the considered studies (\Cref{sec:literature-overview}) include:
\begin{itemize}
    \item Some studies do not clarify how the training-test data split was performed and whether an additional validation set was used for hyper-parameter tuning. Determining this information is particularly challenging in cases where the source code is missing.
    \item Most of the studies that propose new methodologies (e.g. improvements for \gls{dl} architectures or pre/post-processing pipelines) usually rely on different datasets, thus making it difficult to compare the added value of the proposed improvements with respect to other studies and existing methods. Additionally, a detailed ablation study (i.e. a systematic analysis where components of a model are successively removed) is often missing to empirically justify some of the methodological contributions. 
    \item Most studies do not investigate how sensitive the proposed methods are to random initialization or the training-validation split.
    \item There is a risk of inconsistent/biased results in the studies that apply the same single model on the entire \gls{roi} since this implies including the results based on the data on which the model was trained.
\end{itemize}

Based on these limitations and other more general aspects, we provide a few recommendations in the next section that could be adopted in future studies. We then briefly discuss in \Cref{sec:discussion-glacier-evolution} another promising area of glacier-related research where \gls{dl} also starts playing a significant role, i.e. glacier mass balance and evolution modelling. We close with an outlook in \Cref{sec:oulook}.

\subsection{Recommendations}
\label{sec:recommendations}

Addressing the limitations we identified in the studies we discussed in our review could potentially facilitate model inter-comparison and accelerate the progress in the field of \gls{dl}-based glacier mapping. Therefore, we here list a few recommendations, most of them being standard best practices in \gls{ml}-based scientific research\footnote{For a more detailed perspective on this, we recommend, e.g. \citep{heil_reproducibility_2021} and guidelines like REFORM (Reporting Standards For Machine Learning Based Science) \citep{kapoor_reforms_2023}}: 

\begin{itemize}
    \item \textbf{Open Data \& Code.} Publishing both the source code and the (processed) data is crucial for advancing scientific research and promoting transparency and reproducibility. If publishing the data is not possible (e.g. due to proprietary constraints), we strongly encourage to at least publish the output of the models, e.g. as an inventory in the case of glacier extent mapping. Additionally, even if the raw data is often available from the original sources, publishing the ready-to-train data can stimulate new methodological developments and facilitate model inter-comparison. We also encourage re-using existing datasets instead of building new ones from scratch when the main goal is to propose a new methodology and quantify its performance. 
    \item \textbf{Benchmark Datasets.} As an extension to the previous point, we encourage efforts towards building (large-scale) benchmark datasets similar to the prominent ones in \gls{ml}, e.g. ImageNet \citep{deng_imagenet_2009}. Such datasets enable researchers to evaluate, compare, and improve the methods, accelerating progress and innovation. See, e.g. \citet{long_creating_2021} for guidance on how to build a benchmark dataset in Remote Sensing applications. For calving-front detection, such examples exist, e.g. CALFIN \citep{cheng2021_calving} and CaFFe \citep{gourmelon2022_calving}. For glacier extent mapping, the work of \citet{maslov_towards_2024} can be considered a first attempt towards building a large-scale benchmark dataset.
    \item \textbf{Random Initialization/Data Split Sensitivity Analysis.} \gls{dl} models are sensitive to the random weights initialization, with a non-trivial relationship between initialization and final performance \citep{arpit_how_2019}. To account for this effect, one should ideally also evaluate the impact of the training-validation split on the model performance, especially for cases where the weight initialization is fixed (e.g. when using pre-trained models).  
    \item \textbf{Detailed Ablation Studies.} When proposing various methodological improvements to existing architectures, one should always perform ablation studies to evaluate the impact of each improvement, thereby providing empirical evidence for the proposed benefits. Where possible, the same applies to post-processing pipelines. Furthermore, ablation studies can also be performed to study the contribution/importance of the individual data modalities in case multiple are used.
    \item \textbf{Uncertainty Quantification.} Although being an active area of \gls{dl} research \citep{abdar_review_2021, gawlikowski_survey_2023}, some effort should be invested into quantifying the uncertainty in the predictions, e.g. by training an ensemble, and investigating the quality of the uncertainties, which can for instance be realized by checking if the uncertainties agree with the actual errors the model is making.
    \item {\textbf{Cross-Validation with Regional Split.}} When one has to use a model for an entire dataset, e.g. to build a regional inventory, we recommend training multiple models with a regional cross-validation scheme to avoid running the model on the training data. In particular, for studies focused on glacier area change analysis, regional cross-validation can prevent the risk of biased results that are a consequence of memorization (see \cref{sec:glacier-extent-mapping-area-change}).
\end{itemize}

\subsection{Deep Learning for Modelling Glacier Mass Balance and their Evolution}
\label{sec:discussion-glacier-evolution}

\gls{dl} for glacier mass balance (typically relying on regression-based types of approaches) is generally at an earlier stage of development in comparison to the glacier mapping efforts (classification-based types of approaches) on which we focused in this review. The few studies that have so far used \gls{ml} for glacier mass balance problems have mostly relied on more classical (i.e. non-\gls{dl}) models or shallow \gls{mlp} networks. A major issue that these studies face relates to the generally limited data availability for training, while mass balance models would, ideally, need high spatial and temporal resolution observations to be able to capture the complex interaction between climate and topography. For instance, to be able to capture the seasonal mass variability as a response to the local climatic conditions, we would need both accurate meteorological records at the glacier level, especially for precipitation, and also mass balance measurements with a (sub)seasonal frequency. This is rarely the case in practice since only a few hundred glaciers are being monitored with a (sub)annual frequency, and, second, weather stations are very sparse in mountain regions. As a result, we have to rely on geodetic \glspl{mb}, which are usually available only at a multi-annual scale, thus averaging out the intraannual variability driven strongly by the local weather. Additionally, since weather measurements are limited, many studies use reanalysis data, which usually comes at a very coarse resolution, e.g. ERA5 at 0.25$^{\circ}$ resolution (around 27-28 km at the equator) \citep{hersbach_era5_2020}.

Some of the first works on \gls{mb} using \gls{ml} are those of \citet{bolibar_deep_2020_tc, bolibar_deep_2020_essd}, who introduced the ALpine Parameterized Glacier Model (ALPGM\footnote{\url{https://github.com/JordiBolibar/ALPGM}}), a glacier evolution model that uses an \gls{mlp} network with four hidden layers for \gls{mb} estimation. The \gls{mlp} model takes as input 34 topo-climatic predictors, e.g. monthly average temperature and snowfall, mean and max altitude, the slope of the glacier tongue, and it is trained with glacier-wide annual \glspl{mb} for 32 French Alpine glaciers. By combining this \gls{ml}-based \gls{mb} component with a glacier evolution module (i.e. a simple geometry parametrization that models the ice dynamics), the transient evolution of glaciers can be modelled. The model was also applied to infer the evolution of the French Alpine glaciers until the end of the century, using climate models' projections under various scenarios, predicting a volume loss between 75
and 88\% \citep{bolibar_nonlinear_2022}. 

Other \gls{ml} studies have also attempted to model the mass balance component of glaciers, for instance focusing on the winter mass balance \citep{guidicelli_spatio-temporal_2023} using gradient boosting regressor (GBR) or \citet{anilkumar_modelling_2023} who used point mass balances (vs. glacier-wide mass balance in works of \citet{bolibar_deep_2020_tc, bolibar_nonlinear_2022}) through various techniques, i.e. random forest (RF), GBR, support vector machine, and \glspl{mlp}. \citet{diaconu_uncertainty-aware_2024} made use of the Open Global Glacier Model \citep{maussion_open_2019} to reconstruct annual \glspl{mb} and use the resulting dataset to systematically study various uncertainty estimation methods for \gls{ml} models (e.g. the ensemble method), analysing their impact on the quality of the predictions.

Recently, \gls{dl} efforts have also been focused on modelling the ice dynamical processes within glaciers. \citet{jouvet_deep_2022} created the Instructed Glacier Model (IGM), which employs a convolutional neural network architecture to emulate the behaviour of computationally expensive ice flow models that are based on solving Navier-Stokes (NS) equations (typically referred to as 'Full-Stokes' in glaciology). Harnessing its extreme gain in computational costs (about three orders of magnitude compared to original NS calculations), IGM was then used to invert for various key components in glacier evolution modelling, such as ice thickness \citep{jouvet_inversion_2023, cook_committed_2023}. In a recent update, IGM was retrained to not only reproduce 'expensive' simulations performed with NS but also to include more physical constraints \citep{jouvet_ice-flow_2023}. Other noteworthy recent advances for ice flow modelling through \gls{dl} include the use of universal differential equations (UDEs) to model glacier flow {\citep{bolibar_universal_2023}. In the latter work, by combining an ice flow model with differential equations with an embedded neural network, a parameter (i.e. the creep component of the ice flow) can be automatically learnt from data as a nonlinear function. This approach combines the advantages of mathematical models (e.g. interpretable, incorporates domain knowledge) with those of \gls{ml} (e.g. flexible, data-driven), which can also serve as a framework for discovering new empirical laws for glacier processes.

To conclude, \gls{dl} applied for glacier mass balance and evolution modelling is a young but rapidly evolving field that is likely to substantially change future models for glacier projections. Like glacier mapping through \gls{dl}, this field will largely benefit from new data becoming available.

\subsection{Outlook}
\label{sec:oulook}

We see many promising directions in the field of automatic glacier mapping based on \gls{dl}. First, we expect (and encourage) the development of large-scale benchmark datasets that will accelerate the methodological developments. Second, we recommend using time-series input data, as this can potentially mitigate the errors caused by (partial) occlusions in single images, e.g. by clouds. Alternatively, single-image model approaches can be used on multiple acquisitions, from which a (temporal) prediction ensemble can be built. Third, we suggest exploring more sophisticated data fusion techniques instead of simply concatenating the input bands from all the sources, as in most current studies. 

We also expect a rapid development of \gls{dl}-based methods for glacier evolution modelling, briefly discussed here in \Cref{sec:discussion-glacier-evolution}. These developments will largely profit from the increasing number of (global) products in the field, e.g. \citet{hugonnet_accelerated_2021}, with many other promising datasets currently being finalized \citep{dussaillant_glacier_2024, zemp_reconciled_2024}. Additionally, given the complexity of glacier-climate interactions, we advocate for developing methods that can make use of expert knowledge, e.g. by including physical constraints in the models, with the recent work by \citet{bolibar_universal_2023} as an example.

\section{Summary}
\label{sec:summary}

In this work, we provided an overview of the glacier mapping literature based on \gls{dl}, highlighting the methodological contributions, the regions on which each study is focused and the data modalities used for training. We divided the studies into two major sub-fields in glacier mapping, which evolved in parallel: i) glacier extent mapping and ii) calving front detection. Both types of classification problems are usually tackled using multi-source datasets, showing the benefit of using \gls{dl} for automatically extracting the relevant features from each modality. We then provide a compact summary of the available resources (data and source codes) to facilitate further experimentation.

\section{Resources}
\label{sec:resources}

In \Cref{tab:resources}, we compiled a list of general resources useful for glacier-related studies, most of them having been used by the studies discussed in this work. Additionally, in \Cref{tab:resources-papers} we provide the full links to the code, data and generated outputs from the studies that make them publicly available (note that for the Web version of this article, the same information is also provided in \Cref{tab:glacier-extent-mapping-standard,tab:glacier-extent-mapping-area-change,tab:glacier-extent-mapping-rock-glaciers,tab:cf-detection} as hyperlinks). 

\begin{table}[!ht]
    \centering
    \begin{NiceTabular}{|m{0.2\textwidth}|m{0.45\textwidth}|m{0.28\textwidth}|}
    \toprule
    \RowStyle[bold]{}
    
    \Block{1-1}{Name} &
    \Block{1-1}{URL} &
    \Block{1-1}{Comments} \\

    \midrule
    \Block{1-1}{National Snow and Ice Data Center (NSIDC)} & 
    \Block{1-1}{
            \begin{itemize}[leftmargin=0.25cm, itemsep=-0.4ex]
            \item \url{https://nsidc.org/data/search\#keywords=glaciers}
        \end{itemize}
    } &
    \Block{1-1}{large database with many resources} \\

    \midrule
    \Block{1-1}{Global Land Ice Measurements from Space (GLIMS)} & 
    \Block{1-1}{
            \begin{itemize}[leftmargin=0.25cm, itemsep=-0.4ex]
            \item \url{https://www.glims.org/}
            \item \url{https://www.glims.org/maps/glims}
        \end{itemize}
    } &
    \Block{1-1}{large dataset containing series of glacier outlines} \\
    
    \midrule
    \Block{1-1}{Randolph Glacier Inventory (RGI)} & 
    \Block{1-1}{
            \begin{itemize}[leftmargin=0.25cm, itemsep=-0.4ex]
            \item \url{http://www.glims.org/rgi_user_guide/welcome.html}
        \end{itemize}
    } &
    \Block{1-1}{globally complete (static) inventory of glacier outlines } \\

    \midrule
    \Block{1-1}{World Glacier Monitoring Service (WGMS)} & 
    \Block{1-1}{
            \begin{itemize}[leftmargin=0.25cm, itemsep=-0.4ex]
            \item \url{https://wgms.ch/about_wgms/}
        \end{itemize}
    } &
    \Block{1-1}{observations on changes in mass, volume, area and length of glaciers with time (glacier fluctuations), globally} \\

    \midrule
    \Block{1-1}{Glacier Monitoring in Switzerland (GLAMOS)} & 
    \Block{1-1}{
            \begin{itemize}[leftmargin=0.25cm, itemsep=-0.4ex]
            \item \url{https://www.glamos.ch/en}
        \end{itemize}
    } &
    \Block{1-1}{observations on changes in mass, volume, area and length of glaciers with time (glacier fluctuations) in the Swiss Alps} \\

    \midrule
    \Block{1-1}{swisstopo} & 
    \Block{1-1}{
            \begin{itemize}[leftmargin=0.25cm, itemsep=-0.4ex]
            \item \url{https://www.swisstopo.admin.ch/en}
            \item \url{https://s.geo.admin.ch/7mobcus7cy55}
        \end{itemize}
   } &
    \Block{1-1}{Switzerland's geoinformation centre from the Federal Office of Topography} \\
    
    \midrule
    \Block{1-1}{Open Global Glacier Model (OGGM)} & 
    \Block{1-1}{
            \begin{itemize}[leftmargin=0.25cm, itemsep=-0.4ex]
            \item \url{https://docs.oggm.org/en/stable/introduction.html}
            \item \url{https://docs.oggm.org/en/stable/shop.html}
            \item \url{https://oggm.org/tutorials/stable/notebooks/10minutes/machine_learning.html}
        \end{itemize}
   } &
    \Block{1-1}{open source modelling framework \& source of many post-processed datasets} \\

    \bottomrule
    
    \end{NiceTabular}
    \caption{Miscellaneous (data) resources for glacier studies}
    \label{tab:resources}
\end{table}

\begin{table}[!ht]
    \centering
    \begin{NiceTabular}{|m{0.015\textwidth}|m{0.015\textwidth}|m{0.16\textwidth}|m{0.65\textwidth}|}
    \toprule
    \RowStyle[bold]{}
    
    \Block[c]{1-2}{Category} & &
    \Block{1-1}{Publication} &
    \Block{1-1}{Resources} \\
    
    \midrule

    \Block[c]{7-1}<\rotate>{Glacier extent mapping \\ (\Cref{sec:glacier-extent-mapping})} &
    \Block[c]{3-1}<\rotate>{General \\ (\Cref{sec:glacier-extent-mapping-standard})} &
    \Block{1-1}{\citet{chu_glacier_2022}}  & 
    \Block[l]{1-1}{
        C: \url{https://github.com/yiyou101/Attention-DeepLab-V3plus} \\
        D: \url{https://pan.baidu.com/s/1P0FFkq3zrIbYfDVTLC_soA?pwd=ctsa}
    } \\

     \cmidrule{3-4}
     
     &
     &
    \Block{1-1}{\citet{maslov_glavitu_2023}}  & 
    \Block[l]{1-1}{
        C: \url{https://github.com/konstantin-a-maslov/GlaViTU-IGARSS2023} \\
        D: \url{https://drive.google.com/drive/folders/1ivi18vaGXdsaxbJLKAt9R8rzOUvgg1Em}
    } \\

    \cmidrule{3-4}

     &
     &
    \Block{1-1}{\citet{maslov_towards_2024}}  & 
    \Block[l]{1-1}{
        C: \url{https://github.com/konstantin-a-maslov/towards_global_glacier_mapping} \\
        D: \url{https://drive.google.com/drive/folders/1wu_DTCknr5ozu5e8FYqRlMLcmNGbOsxA}
    } \\

    \cmidrule{2-4}

    &
    \Block[c]{2-1}<\rotate>{change analysis \\ (\Cref{sec:glacier-extent-mapping-area-change})} &
    \Block{1-1}{\citet{roberts-pierel_changes_2022}}  & 
    \Block[l]{1-1}{
        O: \url{https://doi.org/10.7265/8esq-w553}
    } \\
    
    \cmidrule{3-4}
    
    &
    &
    \Block{1-1}{\citet{diaconu_detailed_2023}}  & 
    \Block[l]{1-1}{
        C: \url{https://github.com/dcodrut/glacier_mapping_alps_tccml} \\
        D: \url{https://huggingface.co/datasets/dcodrut/glacier_mapping_alps}
    } \\
    
    \cmidrule{2-4}

    &
    \Block[c]{2-1}<\rotate>{Rock glaciers \\ (\Cref{sec:glacier-extent-mapping-rock-glaciers})} &
    \Block{1-1}{\citet{hu_mapping_2023}}  & 
    \Block[l]{1-1}{
        \vspace{0.25cm}
        C: \url{https://zenodo.org/records/8223732} \\
        O: \url{https://doi.pangaea.de/10.1594/PANGAEA.938686}
        \vspace{0.25cm}
    } \\
    
    \cmidrule{3-4}
    
    &
    &
    \Block{1-1}{\citet{sun_tprogi_2024}}  & 
    \Block[l]{1-1}{
        \vspace{0.25cm}
        O: \url{https://zenodo.org/records/10732042}
        \vspace{0.25cm}
    } \\
    
    \midrule

    \Block[c]{10-2}<\rotate>{Calving-front detection \\ (\Cref{sec:cf-detection})} &
    &
    \Block{1-1}{\citet{heidler2022_hedunet}}  & 
    \Block[l]{1-1}{
        C: \url{https://github.com/khdlr/HED-UNet} \\
        D: \url{https://geoservice.dlr.de/web/maps/eoc:icelines}
    } \\

    \cmidrule{3-4}

    &
    &
    \Block{1-1}{\citet{mohajerani2019_detection}}  & 
    \Block[l]{1-1}{
        C/D: \url{https://github.com/yaramohajerani/FrontLearning/}
    } \\

    \cmidrule{3-4}
    
    &
    &
    \Block{1-1}{\citet{zhang2019_automatically}}  & 
    \Block[l]{1-1}{
        O: \url{https://doi.pangaea.de/10.1594/PANGAEA.897066}
    } \\

    \cmidrule{3-4}
    &
    &
    \Block{1-1}{\citet{zhang2021_automated}}  & 
    \Block[l]{1-1}{
        C: \url{https://github.com/enzezhang/FrontDL3} \\ 
        D: \url{https://doi.pangaea.de/10.1594/PANGAEA.923270} \\ 
        O: \url{https://doi.pangaea.de/10.1594/PANGAEA.923272}
    } \\

    \cmidrule{3-4}
    &
    &
    \Block{1-1}{\citet{cheng2021_calving}}  & 
    \Block[l]{1-1}{
        C/D/O: \url{https://github.com/daniel-cheng/CALFIN}
    } \\

    \cmidrule{3-4}
    &
    &
    \Block{1-1}{\citet{heidler2023_deep}}  & 
    \Block[l]{1-1}{
        C/O: \url{https://khdlr.github.io/COBRA/}
    } \\

    \cmidrule{3-4}
    &
    &
    \Block{1-1}{\citet{gourmelon2022_calving}}  & 
    \Block[l]{1-1}{
        C: \url{https://github.com/Nora-Go/Calving_Fronts_and_Where_to_Find_Them} \\ 
        D: \url{https://doi.pangaea.de/10.1594/PANGAEA.940950} \\ 
        O: \url{https://zenodo.org/records/6469519}
    } \\

    \cmidrule{3-4}
    &
    &
    \Block{1-1}{\citet{wu2023_amdhooknet}}  & 
    \Block[l]{1-1}{
        C: \url{https://github.com/RiverNA/AMD-HookNet}
    } \\

    \cmidrule{3-4}
    &
    &
    
    \Block{1-1}{\citet{zhang2023_autoterm}}  & 
    \Block[l]{1-1}{
        C: \url{https://zenodo.org/records/8270875} \\ 
        O: \url{https://zenodo.org/records/7782039}
    } \\

    \cmidrule{3-4}
    &
    &
    \Block{1-1}{\citet{wu2024_contextual}}  & 
    \Block[l]{1-1}{
        C/O: \url{https://github.com/RiverNA/HookFormer}
    } \\

    \bottomrule
    
    \end{NiceTabular}
    \caption{Resources (code (C), data\tablefootnote{This refers to the processed training data, not to the raw one. The latter is usually openly available from the original source.}(D) and output\tablefootnote{In case the study has an output product (e.g. an inventory).} (O)) extracted from the works reviewed in \Cref{sec:glacier-extent-mapping}}
    \label{tab:resources-papers}
\end{table}

\FloatBarrier
\section*{Acknowledgments} 

Codruț-Andrei Diaconu is supported by the Helmholtz Association through the joint research school Munich School for Data Science - MuDS (grant number: HIDSS-0006).
Konrad Heidler was supported by the German Federal Ministry for Economic Affairs and Climate Action in the framework of the "national center of excellence ML4Earth" (grant number: 50EE2201C).
Jonathan Bamber was supported by the European Union's Horizon 2020 research and innovation programme through the project Arctic PASSION (grant number: 101003472) and the German Federal Ministry of Education and Research (BMBF) in the framework of the international future AI lab "AI4EO -- Artificial Intelligence for Earth Observation: Reasoning, Uncertainties, Ethics and Beyond" (grant number: 01DD20001)

\printbibliography

\end{document}